\documentclass{article}

\usepackage[preprint]{corl_2026}

\usepackage{iftex}
\ifPDFTeX
  \usepackage[utf8]{inputenc}
  \usepackage[T1]{fontenc}
  \usepackage{newunicodechar}
  \newunicodechar{α}{$\alpha$}
\else
  \usepackage{fontspec}
  \defaultfontfeatures{Scale=1.06}
\fi
\usepackage{hyperref}
\usepackage{url}
\usepackage{booktabs}
\usepackage{amsfonts}
\usepackage{amsmath}
\usepackage{amssymb}
\usepackage{nicefrac}
\usepackage{sectsty}
\usepackage{setspace}
\usepackage{microtype}
\usepackage{xcolor}
\usepackage{graphicx}
\usepackage{subcaption}
\usepackage{tcolorbox}
\tcbuselibrary{skins,breakable}
\usepackage{longtable}
\usepackage{enumitem}
\usepackage{tikz}

\definecolor{deepgreen}{HTML}{0B5D43}
\definecolor{linkgreen}{HTML}{0A6B4A}
\definecolor{softgreen}{HTML}{DCEBE5}
\definecolor{accent}{HTML}{0B5D43}
\hypersetup{
  colorlinks=true,
  linkcolor=linkgreen,
  citecolor=linkgreen,
  urlcolor=linkgreen,
  filecolor=linkgreen,
  pdftitle={What Makes an Efficient VLA? Navigating Action-Head Design, Scaling, and Latency},
  pdfauthor={Luoyang Sun, Guoyang Xia, Fengfa Li, Lei Ren, Xinyu Cui, Haifeng Zhang, Fangxiang Feng, Kaike Zhang, Kun Zhan, Yan Xie, Jun Wang, Cheng Deng},
  pdfsubject={A controlled, latency-paired study of modular vision-language-action models},
  pdfkeywords={vision-language-action models, robot learning, efficient inference, action-head design}
}
\microtypesetup{protrusion=true}
\AtBeginDocument{\newgeometry{textwidth=6.3in,textheight=9in,top=1in,headheight=12pt,headsep=25pt,footskip=30pt}}
\sectionfont{\color{deepgreen}\sffamily\large\bfseries\raggedright}
\subsectionfont{\color{deepgreen}\sffamily\normalsize\bfseries\raggedright}
\subsubsectionfont{\color{deepgreen}\sffamily\normalsize\bfseries\raggedright}
\paragraphfont{\normalsize\bfseries}
\newtcolorbox{finding}{
  colback=softgreen!45, colframe=deepgreen, boxrule=0.5pt, arc=2.5pt,
  left=8pt, right=8pt, top=6pt, bottom=6pt, enhanced,
  before skip=6pt, after skip=6pt,
  borderline west={2.6pt}{0pt}{deepgreen},
}

\definecolor{lightabstract}{HTML}{F3F5F4}
\renewenvironment{abstract}{%
  \begin{tcolorbox}[enhanced,breakable,colback=lightabstract,colframe=deepgreen,
    boxrule=0.8pt,arc=4pt,left=14pt,right=14pt,top=11pt,bottom=12pt,width=\textwidth,
    before skip=10pt,after skip=12pt]%
  \begingroup\small\setstretch{1.05}%
  \begin{center}{\sffamily\bfseries\large\color{deepgreen}Abstract}\end{center}%
  \vspace{2pt}\par\noindent\ignorespaces
}{%
  \par\endgroup\end{tcolorbox}%
}

\title{What Makes an Efficient VLA? Navigating Action-Head Design, Scaling, and Latency}

\author{%
  Luoyang Sun\textsuperscript{1,2,3,4} \thanks{Equal contribution.} \quad
  Guoyang Xia\textsuperscript{5,4} \footnotemark[1]\quad
  Fengfa Li\textsuperscript{4} \quad
  Lei Ren\textsuperscript{4} \thanks{Project Leader.} \thanks{Corresponding author} \quad
  Xinyu Cui\textsuperscript{1,2} \\
  \textbf{Haifeng Zhang}\textsuperscript{1}  \footnotemark[3]  \quad
  \textbf{Fangxiang Feng}\textsuperscript{5} \quad
  \textbf{Kaike Zhang}\textsuperscript{4} \quad
  \textbf{Kun Zhan}\textsuperscript{4} \quad
  \textbf{Yan Xie}\textsuperscript{4} \\
  \textbf{Jun Wang}\textsuperscript{6} \quad
  \textbf{Cheng Deng}\textsuperscript{7}  \footnotemark[3] \\
  \\
  \textsuperscript{1}Institute of Automation, Chinese Academy of Sciences \\
  \textsuperscript{2}University of Chinese Academy of Sciences \\
  \textsuperscript{3}AI Lab, The Yangtze River Delta \\
  \textsuperscript{4}Li Auto Inc. \\
  \textsuperscript{5}School of Artificial Intelligence, Beijing University of Posts and Telecommunications \\
  \textsuperscript{6}University College London \quad
  \textsuperscript{7}University of Edinburgh
}

\begin{document}
\maketitle

\begin{abstract}
Vision-Language-Action (VLA) models combine a pretrained vision encoder, a language backbone, and an action head, but 
the relative contribution of these design choices has not been established under controlled, latency-paired conditions.
We fix the backbone families (SigLIP2 and Qwen2.5) and the training pipeline, while sweeping action-head design and module scale,
pairing every configuration with measured on-device latency. 
The study yields three findings.
First, 
action-head performance is governed primarily by initialization rather than decoder architecture, loss, or inference budget:
copying the last transformer layers of the language backbone into the head is the single largest lever we measure, at no latency cost, and 
the only axis that consistently improves performance across all module scales tested within the L1 family. Initialization also reorganizes the other axes: flow matching and a heavier decoder pay off only without \textsc{VLM-init}, and
additional inference passes provide no measurable benefit once \textsc{VLM-init} is used,
suggesting that the additional expressiveness compensates
for the compatibility the head lacks.  We read the mechanism as weight transfer: the initialized head remains structurally closer to the backbone in weight space, while a separate attention analysis shows a clearer and more concentrated instruction-attention pattern.  Weight-space CKA does not by itself establish activation-level representation alignment or semantic grounding.  Because we manipulate this state only through initialization, we offer compatibility as the account that best organizes the measurements rather than as a demonstrated cause.  Second, initialization substantially amplifies the return to L1 transformer-head capacity: scaling from $s$ to $l$ gains $9.4$ points with \textsc{VLM-init}, compared with $3.1$ for the matched random-initialized L1 transformer head.  Third, those returns diminish sharply at a model size close to the one today's $\pi$-series VLAs already use, so further growth buys little in-domain accuracy for its added latency.  The three together specify \textsc{EffVLA}, a compact model that matches the strongest open-source VLAs on standard LIBERO, leads on most LIBERO-Plus perturbation axes at lower latency, and transfers to a real SO-ARM101 arm with the recipe unchanged (only the training data is swapped).

\vspace{6pt}\noindent{\color{deepgreen}\textbf{Keywords:}}~Vision-Language-Action model, Robot, Efficient inference

\vspace{6pt}\noindent{\color{deepgreen}\textbf{Code:}}~\href{https://github.com/MindVLA-Team/EFFVLA}{https://github.com/MindVLA-Team/EFFVLA}

\vspace{6pt}\noindent{\color{deepgreen}\textbf{Project Page:}}~\href{https://mindvla-team.github.io/EFFVLA}{https://mindvla-team.github.io/EFFVLA}

\end{abstract}

\section{Introduction}
\label{sec:intro}

A Vision-Language-Action (VLA) model is assembled from three components: a pretrained vision encoder, a pretrained language backbone, and an action head that turns their representations into motor commands. How to assemble them is largely decided by trial and error. $\pi_0$~\citep{black2024pi_0} builds a flow-matching action expert that runs several denoising steps; OpenVLA-OFT~\citep{kim2025fine} replaces it with an MLP regression head reading VLM hidden states; OpenVLA~\citep{kim2024openvla} decodes action tokens autoregressively through the language backbone; FAST~\citep{pertsch2025fast} puts a frequency-domain tokenizer in front of the same idea. Each system fixes a different combination of choices, trains on different data, and reports gains on a different benchmark. 
The field has accumulated improvements without controlled isolation of the contribution of each design choice.
And because a VLA must act in real time, with only tens of milliseconds per action chunk, the latency cost of each choice matters as much as its accuracy, yet it is almost never reported.

\begin{figure}[htbp]
\centering
\includegraphics[width=\linewidth]{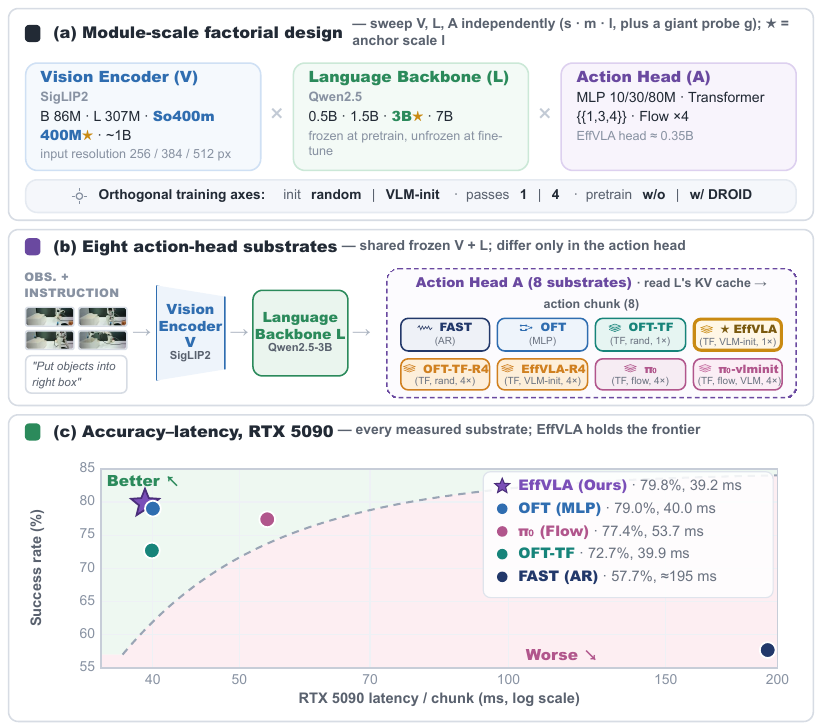}
\caption{\textbf{A controlled design-space study of a modular VLA within one fixed backbone family.} \textbf{(a)}~The module-scale factorial: we hold the SigLIP2-So400m vision encoder ($V$), the Qwen2.5 language backbone ($L$), and the action head ($A$) fixed as an anchor ($V\!=\!L\!=\!A\!=\!l$, star), then sweep each module independently ($s\!<\!m\!<\!l$, plus a larger $g$ probe), alongside action-head design choices: initialization $\in\{$random, \textsc{VLM-init}$\}$, inference passes $\in\{1,4\}$, and DROID pretraining on/off. \textbf{(b)}~The eight action-head substrates share the same $V\!+\!L$ backbone but not one common read interface: \textsc{Fast} decodes through the language trunk, \textsc{OFT} reads VLM hidden states, and the six transformer-block conditions use KV-share. These are \textsc{OFT-TF}, \textsc{OFT-TF-r4}, \textsc{EffVLA} (\textsc{OFT-TF-vlminit}), \textsc{EffVLA-r4} (\textsc{OFT-TF-vlminit-r4}), $\pi_0$, and $\pi_0$-\textsc{vlminit}. \textbf{(c)}~\textsc{EffVLA} is the selected single-pass L1$+$\textsc{VLM-init} recipe; on the measured RTX~5090 accuracy--latency plane, it holds the frontier.}
\label{fig:headline}
\end{figure}

This paper provides that measurement. We hold the backbone fixed (Qwen2.5-3B~\citep{qwen2025qwen25} with SigLIP2-So400m~\citep{tschannen2025siglip}), the pipeline fixed (DROID~\citep{khazatsky2024droid} pretraining, then LIBERO~\citep{liu2023libero} fine-tuning), and the protocol fixed (LIBERO-Plus~\citep{fei2025libero}), and we train $63$ cells that vary only the design under study. The study differs from prior design-space sweeps~\citep{li2024towards,kim2025fine} in two ways. First, every cell is paired with wall-clock latency on the same hardware (NVIDIA RTX 5090, bf16, action chunk size 8). Second, the four action-head axes that prior work varies jointly, decoder architecture, training loss, weight initialization, and number of inference passes, are isolated one at a time, so each contribution can be read without confounds. We analyze the action head first, then fix its best form and scale the three modules independently.

We call a fixed choice of the four action-head axes, independent of module size, a \emph{substrate}. The analysis yields three design rules that compose directly into a recipe, answering in turn \emph{what to build}, \emph{where to scale}, and \emph{how big} to make the model.

\textbf{What to build.} One axis dominates the rest, and it is not an architectural one: how the action head is initialized. Copying the last few transformer layers of the language backbone into the head, rather than training it from scratch, adds $7.1$ points on LIBERO-Plus at no measurable latency cost (more than three times what scaling the language backbone across our full sweep buys), and it pays at every module scale we test within the L1 family. Once the head uses \textsc{VLM-init}, the additional modeling complexity provides no further measurable gain:
flow matching loses $4.4$ points to plain L1, an elaborate decoder is no better than a single transformer block, and extra inference passes add nothing \emph{after \textsc{VLM-init}}. Equivalently, the decoder and loss help only without \textsc{VLM-init}, 
where they partially compensate for the absence of VLM initialization,
while the random-initialized L1 head gains $2.1$ points from four passes. \emph{Rule~1: initialize the head from the backbone's last layers, and keep the initialized model simple with single-pass L1 regression.}

\textbf{Where to scale.} Initialization strongly amplifies the return to action-head capacity, and that changes where added compute is worth spending. Under the initialized head, scaling the action head returns about four success-rate points per millisecond of added latency, against roughly $1$ for the vision encoder and $0.15$ for the language backbone; remove the initialization and the same head-scaling yields about a third as much ($+3.1$ points from $s$ to $l$ under random init versus $+9.4$ under \textsc{VLM-init}). 
the action head therefore offers the highest return on additional capacity,
in the tested initialized substrate. \emph{Rule~2: initialize first, then spend the scaling budget on the action head.}

\textbf{How big.} 
Further scaling provides no measurable benefit beyond a modest size.
Scaling any of the three modules past our largest configuration leaves accuracy flat while latency keeps rising: the returns the first two rules identify are already captured at about $3.75$B parameters, close to the budget the $\pi$-series of VLAs~\citep{black2024pi_0} consistently adopts. \emph{Rule~3: stop there; within our tested regime, further scaling buys little in-domain accuracy for its added latency.}

What this paper contributes is not a new architecture but a controlled account of what governs a modular VLA. Its centerpiece is that the dominant axis is how the action head is initialized, not the decoder, loss, or inference budget the literature has focused on. Two distinct analyses are consistent with a compatibility account: weight-space CKA shows that copied projection-weight structure is retained, while attention measurements show greater mass on instruction tokens. The former is not an activation-level representation measure and does not by itself demonstrate semantic grounding. Because we manipulate compatibility only through initialization, we present it as the reading that best organizes the measurements rather than as an established cause (Section~\ref{sec:limitations}). The two scaling rules and the \textsc{EffVLA} recipe follow from the measurements and are consistent with that account. In summary, this paper makes the following contributions:
\begin{itemize}[leftmargin=1.5em]
\item We conduct a controlled, latency-paired study that isolates the four action-head axes and three module-scale axes one at a time, so that each measured gain is attributable to the design choice that changed rather than to a stronger backbone, more data, or a different benchmark.
\item We identify the action head's weight initialization, copying the backbone's last transformer layers into the head, as the most influential axis in our controlled comparisons: it is positive in every matched L1 configuration we test and reorders the other axes, since the decoder and loss axes reverse sign between the initialized and the randomly initialized head while the inference-budget axis stops paying after initialization. Weight-space CKA and attention provide complementary, correlational evidence for a compatibility account; neither establishes activation-level representation transfer or causality.
\item We show that initialization substantially amplifies the return to action-head capacity: the initialized action head is the highest-return module to scale, while the same scaling on a random-initialized transformer head still helps but buys about a third as much.
\item We show that these returns saturate at a modest, $\pi$-series-sized budget, and distil the three findings into \textsc{EffVLA}, which matches the strongest open-source VLAs on standard LIBERO, leads on most LIBERO-Plus perturbation axes at lower latency, and transfers unchanged to a real SO-ARM101 arm.
\end{itemize}

\section{Related Work}
\label{sec:related}

\paragraph{Vision-Language-Action models.} VLAs produce actions in three ways: decoding them as language tokens through the backbone (RT-1/RT-2~\citep{brohan2022rt,brohan2023rt}, OpenVLA~\citep{kim2024openvla}, FAST~\citep{pertsch2025fast}); generating them with diffusion or flow matching, trading inference passes for expressiveness (Diffusion Policy~\citep{chi2025diffusion}, Octo~\citep{team2024octo}, $\pi_0$~\citep{black2024pi_0}, CogACT~\citep{li2024cogact}, GR00T~N1~\citep{bjorck2025gr00t}, Gemini Robotics~\citep{team2025gemini}); or regressing them with a lightweight head, either directly from visual features (ACT~\citep{zhao2023learning}) or from VLM hidden states (OpenVLA-OFT~\citep{kim2025fine}). Each arrived on its own backbone, data, and benchmark, so the three families have never been compared like-for-like, which Sections~\ref{sec:exploration}--\ref{sec:efficientvla} do.

\paragraph{VLA design studies.} Closest to us, RoboVLMs~\citep{li2024towards} compares backbones and policy-head structures, OpenVLA-OFT~\citep{kim2025fine} ablates loss and decoding on throughput grounds, and TinyVLA~\citep{wen2025tinyvla} targets the small-parameter regime. 
Concurrent design-space studies VLANeXt~\cite{wu2026vlanext} and StarVLA-$\alpha$~\cite{ye2026starvla} likewise revisit VLA design choices on LIBERO/LIBERO-Plus under unified protocols, the former distilling a recipe from a broad set of findings and the latter stripping complexity to isolate a few axes.
Our study is distinguished by doing two things at once that, to our knowledge, no prior or concurrent VLA design study does together: it isolates action-head weight initialization as its own axis, the choice we find shapes the rest of the design (Section~\ref{sec:exploration}), and it pairs every design choice with measured wall-clock latency on a common deployment target.  VLANeXt and StarVLA-$\alpha$ vary many choices jointly and on accuracy alone; our contribution is the controlled, latency-paired isolation rather than a new head.

\paragraph{Pretraining compatibility and cross-module transfer.} \citet{kumar2022fine} show that naive fine-tuning can distort pretrained features when the objective competes with them, and \citet{lee2022surgical} show that fine-tuning a few layers can beat adapting the full network. Copying the language backbone's last layers into the action head gives the new module a compatible parameter starting point and is the highest-return choice in our L1 family. The nominal flow-matching comparison is $-2.2$ points, but it also changes the block type and is therefore not an initialization-only effect (Appendix~\ref{app:implementation}).

\paragraph{Latency-aware design and scaling.}
Hardware-aware architecture design has long optimized models against measured deployment cost rather than FLOPs alone
(EfficientNet~\citep{tan2019efficientnet}, MobileNet~\citep{howard2017mobilenets}, MnasNet~\citep{tan2019mnasnet}).
Representative systems efforts further demonstrate that substantial efficiency gains can come from hardware-conscious execution, including precision-aware acceleration in SPARK~\citep{liu2024spark} and adaptive Transformer inference in ASTER~\citep{liu2025aster}, and roofline-based hardware co-design scaling for on-device LLMs~\citep{sun2026hardware}.
Scaling laws further characterize parameter--data trade-offs in monolithic backbones~\citep{kaplan2020scaling,hoffmann2022training}
and across robot embodiments and data volumes~\citep{o2024open,lin2025data}.
We bring this measurement-driven Pareto framing to modular VLA design (Section~\ref{sec:scaling}),
reporting per-module returns against measured on-device latency as descriptive slopes within our tested range rather than power-law fits.

\section{The Four Action-Head Axes}
\label{sec:exploration}

The action head is where modular VLAs differ most, and where the literature offers the least controlled guidance.  This section isolates the choices that define it.  We describe the design space and protocol, then read off the axis that dominates the rest, the action head's weight initialization, and show that it reorders the preferred value of every other axis.

\paragraph{Design space.} A modular VLA pairs a pretrained vision encoder ($V$) and language backbone ($L$) with an action head ($A$). We fix the pair to SigLIP2~\citep{tschannen2025siglip} and Qwen2.5~\citep{qwen2025qwen25} and organize the action-head study around four design axes (not every value is architecturally applicable to every substrate):
\begin{itemize}[leftmargin=0.8em]
\item \textbf{decoder}: an MLP regression head on VLM hidden states, a stack of transformer blocks reading the backbone through a KV-share interface, or an autoregressive trunk that decodes action tokens through the language backbone itself;
\item \textbf{loss}: L1 regression, flow matching~\citep{lipman2022flow,liu2022flow}, or next-token prediction for the AR trunk;
\item \textbf{initialization} (transformer-block heads only): random, or copied from the last transformer layers of the language backbone (\textsc{VLM-init});
\item \textbf{inference passes}: one forward pass per chunk, or four (denoising steps for flow matching; the same block re-run four times for L1).
\end{itemize}
In the $L_1$ substrates the transformer block is the standard attention-plus-MLP block, so the random and \textsc{VLM-init} conditions share an identical architecture and the initialization contrast varies weights alone. Table~\ref{tab:substrates} names the eight \emph{substrates} (fixed choices of the four axes) we use as shorthand throughout.

\paragraph{Protocol and conventions.} Every cell follows the same two-stage pipeline: action-head pretraining on DROID~\citep{khazatsky2024droid} with public SigLIP2 and Qwen2.5 checkpoints, then full-VLA fine-tuning on LIBERO~\citep{liu2023libero} with all backbones unfrozen. Evaluation is on LIBERO-Plus~\citep{fei2025libero} (four task suites $\times$ seven zero-shot perturbation axes, ${\sim}1000$ rollouts per cell); the metric is success rate. Latency is end-to-end on an NVIDIA RTX~5090 (bf16, batch~$1$, action chunk size~$8$, \texttt{torch.compile}, mean of $100$ runs), the target recent VLA work reports against~\citep{kim2024openvla,black2024pi_0,kim2025fine}. Three conventions hold everywhere and we state them once. First, each per-cell success rate averages ${\sim}1000$ LIBERO-Plus rollouts, giving a binomial standard error of about $1.4$ points. The five-seed \textsc{EffVLA}--\textsc{OFT} check in Appendix~\ref{app:seeds} separately measures training variation: the two models have standard deviations of $1.2$ and $1.5$ points, and \textsc{EffVLA} ranks higher under all five seeds. We therefore treat sub-point gaps as unresolved and gaps of only a few points as suggestive, while using the repeated-seed result to support the stability of that close anchor ordering. The per-axis and per-suite breakdowns (Tables~\ref{tab:per-axis-ablation},~\ref{tab:full-grid-axis}) use fewer rollouts and are less precise. Second, we sweep $V$, $L$, and $A$ over three sizes $s\!<\!m\!<\!l$ (with $l$ the anchor), plus a larger probe $g$ used only to test saturation (Section~\ref{sec:scaling}): $V_{s,m,l,g}=86/307/400\text{M}/{\sim}1\text{B}$ and $L_{s,m,l,g}=0.5/1.5/3/7$B, while the transformer action head grows from $\{1,3,4\}$ blocks (${\sim}0.35$B at $l$) to ${\sim}0.65$B at $g$; the MLP head is the sole exception, scaling over $10/30/80$M ($s/m/l$) with no $g$. We trained $63$ cells (the design space is not a flat product across substrates because the action-head sizes differ per substrate, so we report the actual cell count rather than a full-grid fraction), prioritizing the largest-scale anchor for every substrate and the per-module sweeps used in Sections~\ref{sec:exploration} and~\ref{sec:scaling}; protocol and grid details are in Appendix~\ref{app:implementation} through~\ref{app:results}.

\begin{table}[htbp]
\centering\small
\caption{Action-head ablation matrix at $V = L = A = l$. Eight named combinations of the four design axes, with LIBERO-Plus mean success rate. Our recipe \textsc{EffVLA} is the \textsc{OFT-TF-vlminit} substrate (highlighted).  ``TF block'' denotes a transformer-block stack; per-substrate details are in Appendix~\ref{app:implementation}. }
\label{tab:substrates}
\begin{tabular}{lcccccc}
\toprule
Substrate & Decoder & Loss & Init & Passes & Succ.\ (\%) \\
\midrule
\textsc{Fast}          & AR trunk   & NTP           & None              & many & $57.7$ \\
\textsc{OFT}           & MLP        & L1            & None              & 1    & $79.0$ \\
\textsc{OFT-TF}        & TF block   & L1            & random            & 1    & $72.7$ \\
\textsc{OFT-TF-r4}     & TF block   & L1            & random            & 4    & $74.8$ \\
\textsc{OFT-TF-vlminit} (\textsc{EffVLA})    & TF block   & L1            & \textsc{VLM-init} & 1    & $\mathbf{79.8}$ \\
\textsc{OFT-TF-vlminit-r4} & TF block   & L1            & \textsc{VLM-init} & 4    & $79.6$ \\
$\pi_0$            & TF block   & flow matching & random            & 4    & $77.4$ \\
$\pi_0$-vlminit     & TF block   & flow matching & \textsc{VLM-init} & 4    & $75.2$ \\
\bottomrule
\end{tabular}
\end{table}

Among the factors we isolate, initialization produces the largest and most consistent difference.
Three rows of Table~\ref{tab:substrates} make the point. 
Autoregressive action decoding incurs substantial latency relative to its measured accuracy:
the autoregressive trunk (\textsc{Fast}) trails every parallel-decoding head by at least $15$ points and runs nearly $5\times$ slower. A plain MLP head (\textsc{OFT}) already lands within $1$ point of the best cell (it reads VLM hidden states rather than the backbone's KV cache, so this MLP-vs-transformer contrast also spans the read interface, which we hold fixed only across the four transformer-block substrates; Appendix~\ref{app:limitations}). Replacing that MLP with a randomly initialized transformer stack lowers success by $6.3$ points ($79.0\to72.7$); copying the backbone layers into the same transformer architecture raises the latter result by $7.1$ points. Flow matching, despite running four inference passes through the head, does not reach the best L1 score at this scale. But these contrasts are entangled: once the AR trunk is set aside, the sign of every remaining axis depends on which others are held fixed, so we disentangle them next.

To disentangle them, Table~\ref{tab:factor-effects} reorganizes the pairwise contrasts of Table~\ref{tab:substrates} by the axis each isolates. Once the head is initialized from the backbone, the measured effects are $+0.8$ for a transformer decoder over an MLP, $-0.2$ for a fourth pass, and $-4.4$ for flow matching relative to four-pass L1. We treat the first two as unresolved at the available evaluation resolution and the last as a substantial measured decrease.

Without \textsc{VLM-init}, replacing the MLP with the transformer decoder costs $6.3$ points, whereas flow matching and a fourth pass add $+2.6$ and $+2.1$. Thus the loss axis reverses sign, the decoder contrast goes from a penalty to an unresolved $+0.8$, and the extra-pass gain disappears after initialization. Within the L1 family, initialization is positive in every matched comparison, worth $+4.8$ to $+7.1$ points at the anchor-scale substrates. Under flow matching, however, the nominal initialization condition also changes the block type and is $2.2$ points lower, so it is not a clean initialization-only contrast (Appendix~\ref{app:implementation}).

One explanation covers this pattern. Every axis that reverses adds \emph{expressive capacity} to the head: a generative loss, a second pass through the same blocks, more decoder machinery. Each pays only while the head is initialized at random, and 
once the head is VLM-initialized, these additional mechanisms provide no further measurable gain
and it becomes pure latency. 
These results suggest that the dominant limitation is compatibility with the backbone representation rather than insufficient expressiveness.
A random head must discover how to read the backbone while it also learns to act, and the expressive machinery partially compensates for that.

The same explanation covers the apparent exception in Table~\ref{tab:substrates}, where a head with far less capacity than any transformer stack lands within a point of the best cell at the same latency. The MLP is the one substrate that never has to learn \emph{how} to read the backbone: it consumes the VLM's hidden states directly instead of attending to its KV cache, so no initialization-sensitive attention interface exists and there is no attention pattern to inherit. This direct readout is also why \textsc{VLM-init} does not apply to it. Its price appears in Section~\ref{sec:scaling}: with no depth to grow, it is the one substrate that cannot turn added capacity into accuracy. Section~\ref{sec:mechanism} makes the account measurable.

\begin{table}[htbp]
\centering\small
\caption{Pairwise factor effects derived from Table~\ref{tab:substrates}. Each row reports the success-rate change (percentage points) for the stated variation and the two absolute rates. Individual cells have about $1.4$ points of evaluation standard error; the five-seed anchor check in Appendix~\ref{app:seeds} finds $1.2$--$1.5$ points of run-to-run variation and preserves the \textsc{EffVLA}--\textsc{OFT} ordering in all five seeds. The clean L1 initialization contrasts are positive; the flow-matching comparison also changes block type and is therefore not initialization-only (Appendix~\ref{app:implementation}).}
\label{tab:factor-effects}
\begin{tabular}{l l l c c c}
\toprule
Axis & Variation & Other axes held fixed & From (\%) & To (\%) & $\Delta$ (pts) \\
\midrule
\textbf{Decoder}
  & MLP $\to$ TF block & L1, random init, 1 pass        & 79.0 & 72.7 & $-6.3$ \\
  & MLP $\to$ TF block & L1, \textsc{VLM-init}, 1 pass  & 79.0 & 79.8 & $+0.8$ \\
\midrule
\textbf{Init}
  & random $\to$ \textsc{VLM-init} & TF block, L1, 1 pass    & 72.7 & 79.8 & $\mathbf{+7.1}$ \\
  & random $\to$ \textsc{VLM-init} & TF block, L1, 4 passes  & 74.8 & 79.6 & $+4.8$ \\
  & random $\to$ \textsc{VLM-init} & TF block, FM, 4 passes  & 77.4 & 75.2 & $-2.2$ \\
\midrule
\textbf{Loss}
  & L1 $\to$ FM & TF block, random init, 4 passes          & 74.8 & 77.4 & $+2.6$ \\
  & L1 $\to$ FM & TF block, \textsc{VLM-init}, 4 passes    & 79.6 & 75.2 & $-4.4$ \\
\midrule
\textbf{Compute}
  & 1 pass $\to$ 4 passes & TF block, L1, random init      & 72.7 & 74.8 & $+2.1$ \\
  & 1 pass $\to$ 4 passes & TF block, L1, \textsc{VLM-init}& 79.8 & 79.6 & $-0.2$ \\
\bottomrule
\end{tabular}
\end{table}

\begin{finding}
\textbf{Finding~1, what to build: align the head with the backbone, then keep it simple.} Within the L1 family, initialization is the largest lever we measure, $+7.1$ points at no latency cost, and positive in every initialization-matched pair we report (Appendices~\ref{app:simplerenv} and~\ref{app:pairs}). Once initialized, the expressive machinery stops paying: flow matching $-4.4$, while a fourth pass $-0.2$ and a heavier decoder $+0.8$ are within noise. We scope the rule to single-pass regression:
under flow matching the same weight initialization reduces performance by $2.2$ points
(Table~\ref{tab:factor-effects}). \emph{Initialize the head from the backbone's last layers; keep the rest simple.}
\end{finding}

This reading is operational: three of the four axes have no best value in the abstract, since their preferred setting is induced once initialization is chosen, whereas initialization is decided on its own merits and is the largest lever we measure. Section~\ref{sec:efficientvla} therefore adopts its best L1 setting, \textsc{VLM-init}.

No substrate wins on every axis, and the perturbation axes are where the heads diverge. Table~\ref{tab:per-axis-ablation} breaks the rollouts down two ways: on the four standard task suites the substrates are close, but on the seven perturbation axes the pure MLP head (\textsc{OFT}) is the most robust to sensor noise and lighting, while \textsc{OFT-TF-vlminit} leads on exactly the spatial--language axes a language-initialized head should help (camera viewpoint, language instructions, object layout). The autoregressive trunk (\textsc{Fast}) collapses specifically on the spatially demanding axes (initial state, camera, noise), where decoding actions token-by-token loses the spatial information the parallel heads retain.

\begin{table}[htbp]
\centering\small
\setlength{\tabcolsep}{3.2pt}
\caption{Per-cell LIBERO-Plus success rate (\%) for the eight action-head configurations at $V = L = A = l$, broken down two ways: by the four LIBERO-Plus task suites (\texttt{spatial}, \texttt{object}, \texttt{goal}, \texttt{long}; left of the rule) and by the seven zero-shot perturbation axes (background, robot initial state, camera, language, sensor noise, object layout, lighting; right of the rule). The two breakdowns are independent marginalizations of the same rollouts and need not share a row mean. \textbf{Bold}: column best; \underline{underline}: second-best.}
\label{tab:per-axis-ablation}
\resizebox{\textwidth}{!}{%
\begin{tabular}{l cccc | ccccccc}
\toprule
 & \multicolumn{4}{c|}{Task suites} & \multicolumn{7}{c}{Perturbation axes} \\
\cmidrule(lr){2-5}\cmidrule(lr){6-12}
Config & Spatial & Object & Goal & Long & BG & Init & Cam & Lang & Noise & Layout & Light \\
\midrule
\textsc{Fast}              & $66.4$ & $66.0$ & $48.2$ & $50.2$ & $87.8$ & $33.3$ & $27.9$ & $75.6$ & $35.9$ & $75.7$ & $88.7$ \\
\midrule
\textsc{OFT}               & \underline{82.8} & $80.1$ & $\mathbf{79.0}$ & \underline{74.0} & \underline{97.8} & \underline{68.0} & $53.8$ & $85.3$ & $\mathbf{79.4}$ & $82.9$ & \underline{98.3} \\
\midrule
\textsc{OFT-TF}            & $78.8$ & $71.4$ & $76.0$ & $64.4$ & $97.0$ & $57.8$ & $53.0$ & $86.9$ & $41.8$ & $62.5$ & $74.0$ \\
\textsc{OFT-TF-r4}         & $81.8$ & $75.4$ & $71.7$ & $70.3$ & $97.7$ & $60.7$ & $55.8$ & \underline{88.5} & $56.4$ & $81.8$ & $\mathbf{99.0}$ \\
\textsc{OFT-TF-vlminit} (\textsc{EffVLA})        & $\mathbf{84.4}$ & $\mathbf{84.5}$ & $75.5$ & $\mathbf{74.7}$ & $96.7$ & $\mathbf{69.8}$ & $\mathbf{68.3}$ & $87.5$ & $66.1$ & $84.8$ & $97.4$ \\
\textsc{OFT-TF-vlminit-r4}     & $\mathbf{84.4}$ & \underline{83.7} & \underline{76.9} & $73.3$ & $\mathbf{98.0}$ & $67.2$ & \underline{64.9} & $\mathbf{89.1}$ & \underline{68.2} & \underline{85.1} & $97.4$ \\
$\pi_0$                & $83.6$ & $\underline{83.7}$ & $72.6$ & $69.7$ & $97.1$ & $67.4$ & $62.9$ & $84.6$ & $64.7$ & $82.1$ & $96.4$ \\
$\pi_0$-vlminit         & $81.6$ & $76.7$ & $73.4$ & $68.9$ & $96.8$ & $66.8$ & $54.7$ & $85.9$ & $53.3$ & $\mathbf{87.0}$ & $97.7$ \\
\bottomrule
\end{tabular}}
\end{table}

\section{Compatibility Evidence}
\label{sec:mechanism}

\begin{figure}[htbp]
\centering
\includegraphics[width=\linewidth]{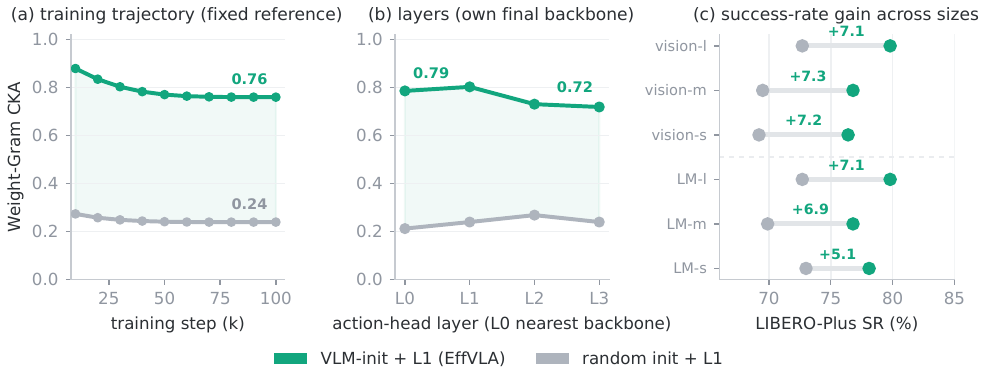}
\caption{\textbf{VLM initialization preserves more backbone-like projection-weight structure, and the initialized models are more accurate.} \textbf{(a)}~Weight-Gram linear CKA to a fixed pretrained-backbone reference during fine-tuning at $V\!=\!L\!=\!A\!=\!l$: the initialized head decreases from about $0.88$ to $0.76$, versus about $0.27$ to $0.24$ for random initialization. \textbf{(b)}~Layerwise Weight-Gram CKA to each model's own final fine-tuned backbone ranges from $0.79$--$0.81$ to $0.72$ under \textsc{VLM-init}, versus $0.21$--$0.27$ under random initialization. These panels measure weights, not input-conditioned activations. \textbf{(c)}~LIBERO-Plus success rates for matched initialization pairs across vision and language sizes (Table~\ref{tab:init-pairs}). The \texttt{vision-l} and \texttt{LM-l} rows reuse the same $l/l/l$ anchor; hence six plotted rows correspond to five unique configurations.}
\label{fig:cka3}
\end{figure}

VLM initialization preserves substantially greater \emph{weight-space structural similarity} to the language backbone than random initialization. This is Weight-Gram CKA over projection matrices, not activation CKA over a common set of inputs; it therefore supports preservation of pretrained weight structure but does not by itself establish similar representations or semantic grounding.
On the $V\!=\!L\!=\!A\!=\!l$ cells, the initialized trajectory decreases from about $0.88$ to $0.76$, while the random-initialized trajectory decreases from about $0.27$ to $0.24$ (Figure~\ref{fig:cka3}a). Across action-head layers, the initialized values vary from roughly $0.79$--$0.81$ at L0--L1 to $0.73$--$0.72$ at L2--L3; the random condition ranges from about $0.21$ to $0.27$ (Figure~\ref{fig:cka3}b). Separately, the success-rate advantage is positive for each one-axis vision and language scale comparison in Figure~\ref{fig:cka3}c (Table~\ref{tab:init-pairs}). The two $l/l/l$ rows are the same shared anchor, so the six plotted rows represent five unique configurations.

\begin{figure}[htbp]
\centering
\includegraphics[width=0.92\linewidth]{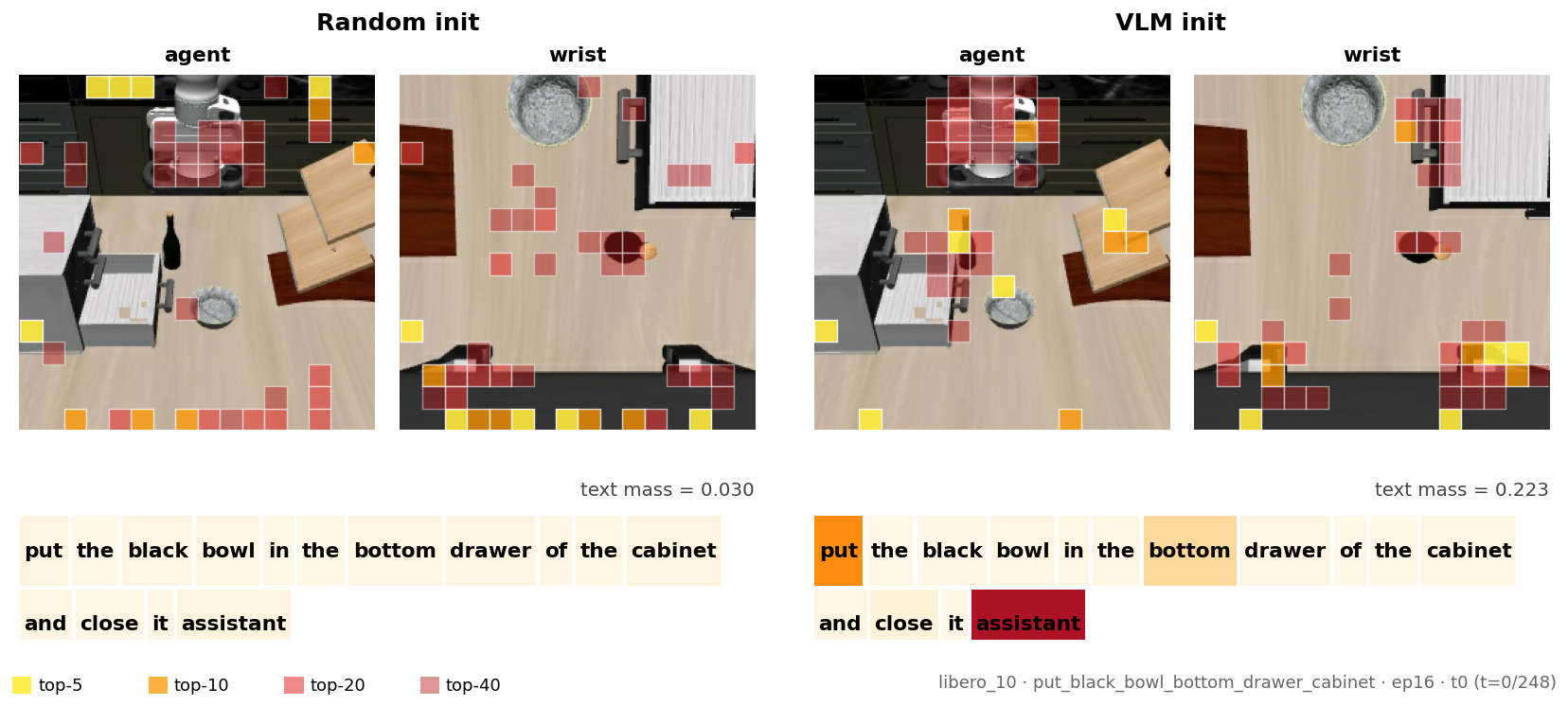}
\caption{\textbf{A single rollout: more concentrated instruction attention under \textsc{VLM-init}.} Task: ``put the black bowl in the bottom drawer of the cabinet and close it.'' Left: random init; right: \textsc{VLM-init}. The vision panels overlay the top-$K$ attended patches (agent and wrist views); the ribbon shades text tokens by attention. The initialized head shows a clearer concentration pattern, greater text mass ($0.223$ vs.\ $0.030$), and visual attention overlapping the bowl. Its strongest text tokens are \texttt{assistant} and \emph{put}; \texttt{assistant} may act as an attention sink or prompt-boundary marker, illustrating that learned attention need not reproduce human noun-centered saliency.}
\label{fig:attention-case}
\end{figure}

The attention analysis supplies distinct behavioural evidence. In the illustrated rollout, the initialized head assigns more total mass to the instruction text ($0.223$ vs.\ $0.030$), with a clearer concentration pattern, and its vision attention overlaps the named bowl. The strongest text tokens are \texttt{assistant} and \emph{put}, rather than \emph{black} or \emph{bowl} (Figure~\ref{fig:attention-case}); the role token may act as an attention sink or prompt-boundary marker. Thus the learned pattern need not match human noun-centered saliency to be structured and task-relevant. Across $128$ rollouts, the initialized head places about four times as much mass on the instruction text, and the aggregate word analysis highlights task-defining object and location nouns. These observations are consistent with, but do not prove, the compatibility account.

\begin{figure}[htbp]
\centering
\includegraphics[width=\linewidth]{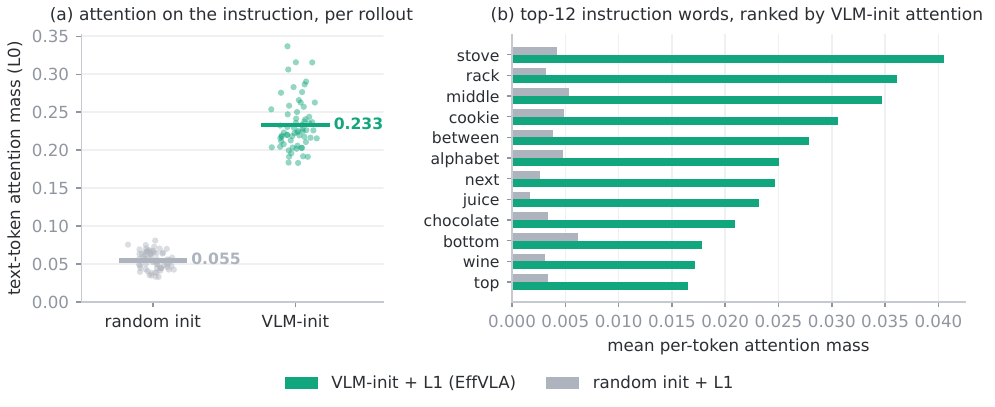}
\caption{\textbf{Instruction attention, aggregated over all rollouts.} \textbf{(a)}~Total attention mass the action head's first layer places on the instruction tokens, one point per rollout ($64$ per condition); bars mark the means, and the two conditions do not overlap. \textbf{(b)}~The top-$12$ instruction words ranked by \textsc{VLM-init} attention are the object and location nouns that name the task, each receiving markedly more mass than under random initialisation. Chat-template role tokens are not instruction words and are excluded from the ranking (Appendix~\ref{app:attention}).}
\label{fig:attention-summary}
\end{figure}

Both readings hold in aggregate, not just on the rollout of Figure~\ref{fig:attention-case}. Over all $128$ rollouts ($64$ per condition, spanning the four task suites) the \textsc{VLM-init} head places roughly four times as much attention mass on the instruction as the random head (per-rollout mean $0.233$ vs.\ $0.055$), and the two distributions do not overlap on a single rollout (Figure~\ref{fig:attention-summary}a). Ranking instruction words by the attention the initialized head gives them, the top of the list is the object and location nouns that name the task (\emph{stove}, \emph{rack}, \emph{cookie}, \emph{alphabet}, \emph{juice}), each receiving several times the mass the random head assigns it (Figure~\ref{fig:attention-summary}b). Thus the aggregate word-level analysis, unlike the single example, is consistent with attention to goal-specifying content. Appendix~\ref{app:attention} reports further rollouts and the token-level detail.

\section{Module Scaling and Saturation}
\label{sec:scaling}

\begin{figure}[htbp]
\centering
\includegraphics[width=\linewidth]{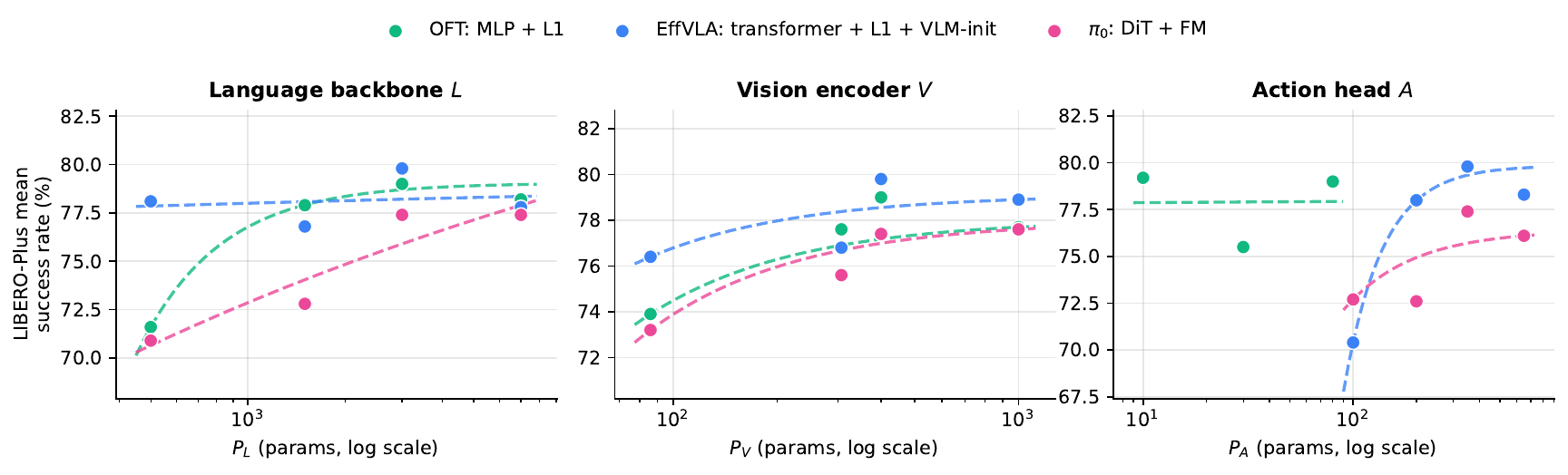}
\caption{\textbf{VLM initialization amplifies the return to action-head capacity, and the measured return saturates.} LIBERO-Plus mean success rate vs.\ module parameter count, one panel per module axis (left to right: $L$, $V$, $A$), with the two non-swept modules held at $l$. The vision and language axes scale at similar rates across the three plotted substrates; the action head separates them: the \textsc{OFT-TF-vlminit} curve rises by $9.4$ points from $s$ to $l$, while \textsc{OFT} is nearly flat and the random-initialized flow-matching $\pi_0$ capacity sweep rises by $4.7$. The matched random-initialized L1 transformer sweep, not plotted here, rises by $3.1$ points (Table~\ref{tab:init-pairs}). These are capacity-scaling effects; separately, at the $l/l/l$ anchor, applying \textsc{VLM-init} to $\pi_0$ changes $77.4$ to $75.2$ ($-2.2$ points; Table~\ref{tab:factor-effects}). Beyond $l$ every plotted curve flattens or dips (rightmost points: $V,L,A\!=\!g$), the empirical basis for fixing the model at $l$ (Finding~3). Substrates: \textsc{OFT} (teal, MLP$+$L1), \textsc{OFT-TF-vlminit} (blue, our \textsc{EffVLA}), $\pi_0$ (pink, flow matching); dashed lines connect measured scales.}
\label{fig:module-scaling}
\end{figure}

\begin{figure}[htbp]
\centering
\includegraphics[width=0.5\linewidth]{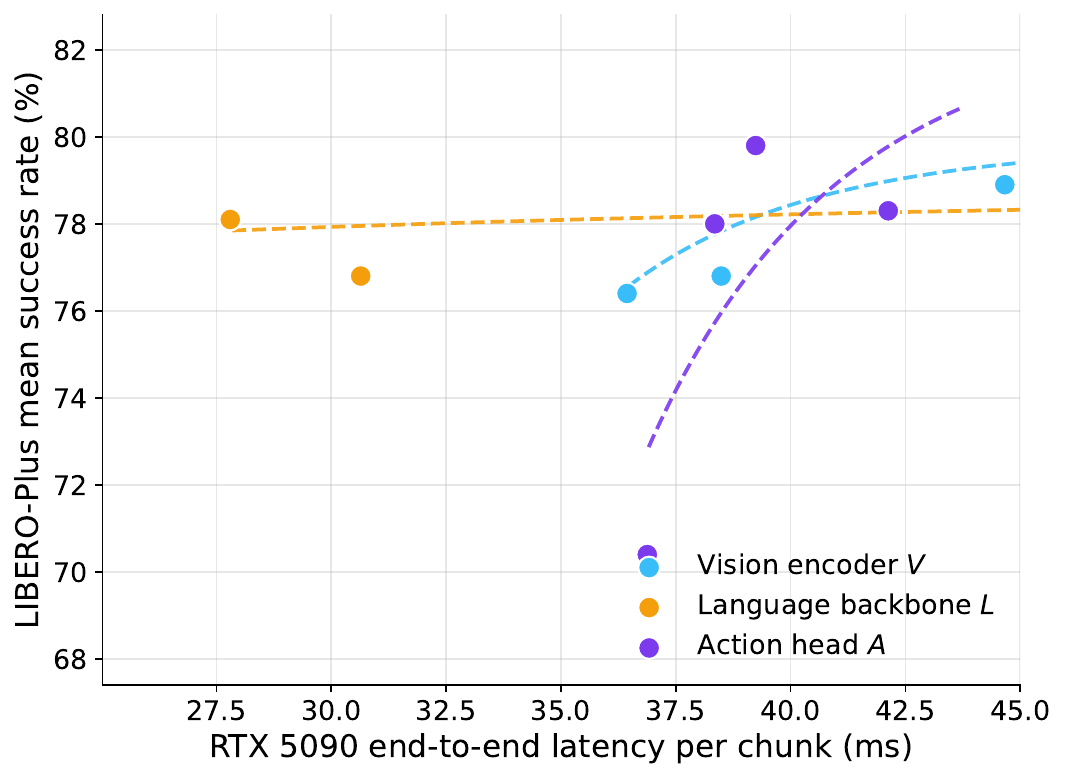}
\caption{\textbf{Where to spend a latency budget under \textsc{OFT-TF-vlminit}.} The three module sweeps of our recipe's substrate on the accuracy--latency plane (success rate vs.\ measured RTX~5090 latency); each module ranges over $\{s, m, l\}$ with the other two at $l$. The action head (violet) has the steepest trade-off, $2.4$~ms for $9.4$ points, against $3.4$ points for the vision encoder (sky) and under $2$ for the language backbone (amber). The dashed saturating fits are extended past the measured points to make the trend visible: every curve plateaus, the latency-side complement to Finding~3.}
\label{fig:module-budget}
\end{figure}

With the substrate fixed, where should extra compute go? We ask which module's \emph{capacity} improves accuracy (Figure~\ref{fig:module-scaling}) and where a unit of \emph{latency} is best spent (Figure~\ref{fig:module-budget}), sweeping $V$, $L$, $A$ at $\{s,m,l\}$ for the three deployment-relevant substrates (\textsc{OFT}, \textsc{OFT-TF-vlminit}, $\pi_0$; the Pareto-dominated random-init control is omitted). Both answers depend on the substrate and point to the same operating size; Appendix~\ref{app:scaling} gives the endpoint deltas and their ratio $\rho$ (success points per ms).

Consider capacity first. Extra action-head parameters help most under \textsc{VLM-init}: the vision encoder and language backbone trace similar saturating curves across substrates, while the largest substrate separation appears on the action-head axis (Figure~\ref{fig:module-scaling}, right). Scaling $A$ from $s$ to $l$ adds $9.4$ points under \textsc{OFT-TF-vlminit} and $3.1$ in the matched random-initialized L1 transformer sweep (Table~\ref{tab:init-pairs}); the random-initialized flow-matching $\pi_0$ capacity sweep separately gains $4.7$ points, while the \textsc{OFT} MLP curve is nearly flat. The $+4.7$ is a size-scaling effect within $\pi_0$, not an initialization gain: at the $l/l/l$ anchor, changing $\pi_0$ from random initialization to \textsc{VLM-init} instead changes $77.4$ to $75.2$ ($-2.2$ points; Table~\ref{tab:factor-effects}). This is the scaling-time counterpart of Finding~1: initialization substantially amplifies, rather than creates, the return to the L1 transformer head's capacity; 
under random initialization, the same capacity yields substantially smaller accuracy gains,
because it must spend that capacity relearning the interface first.

Capacity is not the same as cost, so we turn to latency. On the accuracy--latency plane (Figure~\ref{fig:module-budget}) the action head has the steepest trade-off under our recipe, $2.4$~ms for $9.4$ points, against $3.4$ for the vision encoder and under $2$ for the language backbone; the ordering $A\!>\!V\!>\!L$ holds across the shared latency span without extrapolating a rate. Because the substrate runs a single forward pass, latency rises monotonically with scale, free of the $\pi_0$ depth--width inversion (Appendix~\ref{app:scaling}).

\begin{finding}
\textbf{Finding~2, where to scale: initialize first, then grow the action head.} Action-head capacity provides substantially larger gains after VLM initialization. Under \textsc{VLM-init} the action head is the highest-return module, returning roughly $4$ success points per added millisecond against $\sim\!1$ for the vision encoder and $\sim\!0.15$ for the language backbone. Remove the initialization and the same head-scaling still helps, but buys about a third as much ($+3.1$ points from $s$ to $l$ versus $+9.4$ under \textsc{VLM-init}). \emph{In the tested initialized substrate, the action head provides the largest measured accuracy gain per unit latency.}
\end{finding}

These sweeps deliberately stop at $l$, because pushing past it adds nothing. Under \textsc{OFT-TF-vlminit}, enlarging the action head from $l$ ($350$M) to $g$ ($650$M) moves LIBERO-Plus success rate from $79.8\%$ to $78.3\%$; a $1$B vision encoder scores $78.9\%$ and a $7$B language backbone $77.8\%$, all within noise of or below $l$ (Figure~\ref{fig:module-scaling}, rightmost; Appendix~\ref{app:scaling}). Accuracy has saturated while latency would only keep climbing, so $l$ sits at the efficiency--accuracy knee, close to the budget the $\pi$-series of VLAs~\citep{black2024pi_0} consistently adopts. In this model family and benchmark, our measurements place the saturation point at roughly the same scale. \textsc{EffVLA} (Section~\ref{sec:efficientvla}) is fixed at this scale.

\begin{finding}
\textbf{Finding~3, how big: stop at a modest size.} On LIBERO-Plus, scaling any module past the $l$ configuration leaves accuracy flat while latency keeps climbing, so the ${\sim}3.75$B budget, close to the $\pi$-series scale, is the efficiency/accuracy knee \emph{on this benchmark}. \emph{The $l$ configuration is therefore our preferred operating point; within the tested regime, larger modules add latency without measurable accuracy gains.}
\end{finding}

One ordering follows from all of this: the substrate must be chosen before module sizes, since it decides whether action-head capacity is worth buying at all. 
For the \textsc{VLM-init} substrate, the action head provides the steepest measured accuracy--latency trade-off,
under any other it shifts to vision then language, and beyond $l$ none of them buys much. The per-axis breakdown of which perturbations each module helps is in Appendix~\ref{app:results} (Table~\ref{tab:full-grid-axis}).

\section{EffVLA: The Recipe the Rules Select}
\label{sec:efficientvla}

\paragraph{The recipe.} The analysis specifies one model.  \textsc{EffVLA} is a $4$-layer transformer-block action head copied from the last $4$ transformer layers of Qwen2.5-3B and trained with single-pass L1 regression (Finding~1), reading the backbone through a KV-share interface on a SigLIP2-So400m vision encoder at $256$-pixel input. The action head contributes ${\sim}0.35$B of the model's ${\sim}3.75$B parameters, reflecting the allocation favored by the module-scaling study (Findings~2 and~3). End-to-end latency is $39.2$~ms per action chunk on an RTX~5090, and nothing in the architecture is tuned beyond what the three rules select.  Training follows the two-stage pipeline of Section~\ref{sec:exploration}; full hyperparameters are in Appendix~\ref{app:implementation}.

\paragraph{Open-source baselines on LIBERO and LIBERO-Plus.}  \emph{On the saturated benchmark \textsc{EffVLA} matches the strongest published VLA; on the robustness benchmark it leads on most perturbation axes, and by the widest margins exactly where the language-initialized head is built to help.}  Standard LIBERO has compressed the top tier into a $96$--$99\%$ band, and \textsc{EffVLA} sits inside it at $98.2\%$ (Table~\ref{tab:open-source}, baselines drawn from~\citet{yang2026abot}), $0.4$ points behind \textsc{ABot-M0} and ahead of \textsc{OpenVLA-OFT}.  LIBERO-Plus separates the same models: \textsc{EffVLA} (mean $79.8\%$) leads on six of the seven perturbation axes, with its largest margins on the spatial and language axes that the language-initialized head was built to help, in particular camera viewpoint ($+7.9$ points over \textsc{ABot-M0}) and robot initial state ($+37.9$ over \textsc{OpenVLA-OFT}).  It trails only on sensor noise, where the simpler MLP head was already strongest in our own ablation (Table~\ref{tab:per-axis-ablation}).
This external comparison is consistent with the within-study ablation.

\begin{table}[htbp]
\centering
\small
\setlength{\tabcolsep}{4pt}
\caption{Open-source baseline comparison, in the two-breakdown layout of Table~\ref{tab:per-axis-ablation}: standard-LIBERO success rate by the four task suites (left) and LIBERO-Plus zero-shot success rate by the seven perturbation axes (right). The scores for every method other than \textsc{EffVLA} are those reported by~\citet{yang2026abot} (its Tables~3 and~4), which lists LIBERO-Plus by perturbation axis rather than by task suite; \textsc{EffVLA} (the \textsc{OFT-TF-vlminit} substrate of Sec.~\ref{sec:exploration} at $V\!=\!L\!=\!A\!=\!l$) is the only row we evaluate ourselves, under the same protocol. \textbf{Bold}: column best; \underline{underline}: second-best.}
\label{tab:open-source}
\resizebox{\textwidth}{!}{%
\begin{tabular}{l ccccc | ccccccc}
\toprule
 & \multicolumn{5}{c|}{LIBERO} & \multicolumn{7}{c}{LIBERO-Plus} \\
\cmidrule(lr){2-6}\cmidrule(lr){7-13}
Method & Spatial & Object & Goal & Long & Avg & BG & Init & Cam & Lang & Noise & Layout & Light \\
\midrule
OpenVLA~\citep{kim2024openvla}       & $84.7$ & $88.4$ & $79.2$ & $53.7$ & $76.5$ & $34.8$ & $3.5$ & $0.8$ & $23.0$ & $15.2$ & $28.5$ & $8.1$ \\
$\pi_0$~\citep{black2024pi_0}         & $98.0$ & $96.8$ & $94.4$ & $88.4$ & $94.4$ & $81.4$ & $6.0$ & $13.8$ & $58.8$ & $\underline{79.0}$ & $68.9$ & $85.0$ \\
$\pi_0$-FAST~\citep{pertsch2025fast} & $96.4$ & $96.8$ & $88.6$ & $60.2$ & $85.5$ & $73.2$ & $21.6$ & $\underline{65.1}$ & $61.0$ & $74.4$ & $68.8$ & $73.2$ \\
RIPT-VLA~\citep{tan2025interactive}             & $\underline{98.6}$ & $98.6$ & $\mathbf{99.0}$ & $93.8$ & $97.5$ & $91.6$ & $31.2$ & $55.2$ & $77.6$ & $73.5$ & $74.2$ & $88.4$ \\
OpenVLA-OFT~\citep{kim2025fine} & $97.6$ & $98.4$ & $\underline{97.9}$ & $94.5$ & $97.1$ & $\underline{93.3}$ & $31.9$ & $56.4$ & $79.5$ & $75.8$ & $74.2$ & $88.7$ \\
ABot-M0~\citep{yang2026abot}           & $\mathbf{98.8}$ & $\mathbf{99.8}$ & $\mathbf{99.0}$ & $\mathbf{96.6}$ & $\mathbf{98.6}$ & $91.6$ & $\underline{67.9}$ & $60.4$ & $\underline{86.4}$ & $\mathbf{86.4}$ & $\underline{82.6}$ & $\underline{96.2}$ \\
\textbf{\textsc{EffVLA} (ours)}      & $\underline{98.6}$ & $\underline{99.2}$ & $\mathbf{99.0}$ & $\underline{96.0}$ & $\underline{98.2}$ & $\mathbf{96.7}$ & $\mathbf{69.8}$ & $\mathbf{68.3}$ & $\mathbf{87.5}$ & $66.1$ & $\mathbf{84.8}$ & $\mathbf{97.4}$ \\
\bottomrule
\end{tabular}}
\end{table}

\begin{figure}[htbp]
\centering
\resizebox{\textwidth}{!}{%
\begin{tikzpicture}
 
\def\fw{2.6}
\def\fh{2.0}
\def\gap{0.22}
\def\pad{0.35}
\def\sph{0.13}
\def\spw{0.18}
\def\spgap{0.08}
\def\spspacing{0.38}
 
\pgfmathsetmacro{\contentW}{5*\fw + 4*\gap}
\pgfmathsetmacro{\totalW}{\contentW + 2*\pad}
\pgfmathsetmacro{\stripTop}{\fh + \pad + \spgap + \sph + 0.05}
\pgfmathsetmacro{\stripBot}{-\pad - \spgap - \sph - 0.05}
 
\fill[black!88, rounded corners=2pt]
  (0, \stripBot) rectangle (\totalW, \stripTop);
 
\pgfmathsetmacro{\nholes}{floor(\totalW / \spspacing)}
\foreach \i in {0,...,\nholes} {
  \pgfmathsetmacro{\sx}{\spspacing/2 + \i*\spspacing}
  \pgfmathsetmacro{\chk}{\sx + \spw/2}
  \ifdim\chk pt<\totalW pt
    \fill[black!50, rounded corners=0.8pt]
      (\sx-\spw/2, \fh+\pad+\spgap)
      rectangle
      (\sx+\spw/2, \fh+\pad+\spgap+\sph);
    \fill[black!50, rounded corners=0.8pt]
      (\sx-\spw/2, -\pad-\spgap-\sph)
      rectangle
      (\sx+\spw/2, -\pad-\spgap);
  \fi
}
 
\foreach \i in {0,...,4} {
  \pgfmathsetmacro{\x}{\pad + \i*(\fw+\gap)}
  \node[inner sep=0pt, clip, rounded corners=1pt] at (\x+\fw/2, \fh/2)
    {\includegraphics[width=\fw cm, height=\fh cm]{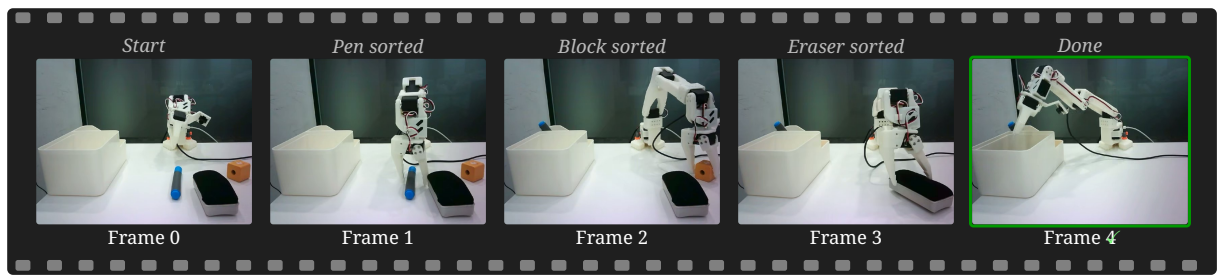}};
}
 
\pgfmathsetmacro{\lastx}{\pad + 4*(\fw+\gap)}
\draw[green!60!black, line width=1pt, rounded corners=1pt]
  (\lastx-0.02, -0.02) rectangle (\lastx+\fw+0.02, \fh+0.02);
 
\foreach \i/\tlab in {0/{Frame 0}, 1/{Frame 1}, 2/{Frame 2}, 3/{Frame 3}, 4/{Frame 4}} {
  \pgfmathsetmacro{\lx}{\pad + \i*(\fw+\gap) + \fw/2}
  \node[white, font=\tiny, inner sep=0pt] at (\lx, -\pad*0.45) {\tlab};
}
\pgfmathsetmacro{\lastlx}{\pad + 4*(\fw+\gap) + \fw/2}
\node[green!50!white, font=\tiny, inner sep=0pt, anchor=west] at (\lastlx+0.3, -\pad*0.45) {\checkmark};

\foreach \i/\slab in {0/Start, 1/{Pen sorted}, 2/{Block sorted}, 3/{Eraser sorted}, 4/Done} {
  \pgfmathsetmacro{\lx}{\pad + \i*(\fw+\gap) + \fw/2}
  \node[white, font=\tiny\itshape, inner sep=0pt, opacity=0.7] at (\lx, \fh+\pad*0.45) {\slab};
}
 
\end{tikzpicture}%
}
\caption{\textbf{Real-robot deployment on SO-ARM101.}  Keyframes of \textsc{EffVLA} executing a sorting task on the SO-ARM101 open-source 6-DOF arm: a pen, a building block, and a board eraser scattered on the table are picked up and placed into their respective compartments of a storage box.  The green border on the last frame marks successful completion.  Stage~A action-head pretraining uses the SO-ARM101 community corpus; Stage~B fine-tunes on a small set of teleoperated demonstrations of the five tasks, merged.  Full rollout videos are in the supplementary material.}
\label{fig:real_robot}
\end{figure}

\paragraph{Real-robot deployment on SO-ARM101.}  The recipe transfers, unchanged, to a physical robot.  We instantiate \textsc{EffVLA} on the low-cost SO-ARM101 open-source 6-DOF arm using the same single-pass L1 head copied from Qwen2.5-3B, with no axis of Section~\ref{sec:exploration} retuned for the new embodiment; only the data pipeline changes, with Stage~A run on the SO-ARM101 community corpus in place of DROID and Stage~B on teleoperated demonstrations of the five tasks, merged and trained jointly so that one model serves all of them.  Across five language-specified sorting tasks spanning one, two, and three objects handled in sequence, it succeeds on $40$ of $50$ trials, with success falling from $10/10$ on the single-object tasks to $5/10$ on the three-object one.  Figure~\ref{fig:real_robot} shows a representative rollout, and Appendix~\ref{app:realrobot} gives the task suite and per-task results.  We report this as a feasibility and transfer result rather than a comparative one: we do not run the alternative action heads on hardware, so these trials show that the LIBERO-selected recipe carries over to a new embodiment, not that it remains the preferred choice there.

\paragraph{Why we stop at $l$, and what smaller variants cost.}  The recipe does not have a larger version.  Finding~3 already established that scaling any module past $l$ leaves success rate flat while latency keeps climbing, 
so a larger \textsc{EffVLA} provides no measurable benefit.
If a deployment instead needs a smaller or simpler model, 
reducing action-head capacity offers the most favorable accuracy--latency trade-off:
a smaller head walks the recipe down the accuracy and latency curve of Figure~\ref{fig:module-budget}, and replacing the transformer stack with a single MLP regression layer (the \textsc{OFT} substrate) costs only about $1$ point at the same latency.  As a side benefit of the $L_1$ family, \textsc{EffVLA} also tolerates skipping the DROID pretraining stage: at $V\!=\!L\!=\!A\!=\!l$ it loses only ${\sim}2$ points without it, against ${\sim}22$ for the flow-matching substrate (Appendix~\ref{app:pretrain}), which is what makes the brief Stage~B sufficient for the SO-ARM101 deployment above.  The per-axis cost of any other single deviation from the recipe is in Appendix~\ref{app:deviations}.

\section{Conclusion}
\label{sec:conclusion}

We presented a controlled, latency-paired study of modular
vision-language-action models. Across the tested configurations, action-head
initialization is more influential than additional decoder complexity:
VLM-initialized heads consistently outperform randomly initialized ones within the L1 comparisons,
while flow matching and repeated inference provide little additional benefit
once this initialization is used. Weight-space CKA shows that copied projection
structure is retained, while a separate attention analysis shows greater
instruction-token mass; together they are consistent with, but do not prove,
a backbone-compatibility account.
Initialization also changes where scaling is most effective. After VLM
initialization, the action head provides the largest measured accuracy gain per
unit latency, with returns saturating near the scale used by EFFVLA. These
results suggest that efficient VLA design should prioritize compatible
initialization and allocate capacity according to measured module-level
accuracy-latency returns.

\section{Limitations}
\label{sec:limitations}

The four most consequential limitations of our study are the following. \emph{Benchmark and failure mode.}  LIBERO-Plus is short-horizon simulation, and we already observe a relevant failure within it: the plain MLP head beats \textsc{EffVLA} on sensor noise (Table~\ref{tab:per-axis-ablation}).  Finding~3's saturation past $l$ may therefore not hold on harder, longer-horizon, or higher-precision tasks where larger backbones could repay (e.g.\ RoboCasa, real long-horizon mobile manipulation).  \emph{Statistical resolution.}  The sweep trains one seed per cell; the five-seed \textsc{EffVLA}--\textsc{OFT} check shows $1.2$--$1.5$ points of run-to-run variation while preserving the same ordering in all five runs (Appendix~\ref{app:seeds}). This supports the stability of that close comparison, while the remaining sweep cells are single-seed. The initialization advantage also recurs across all ten matched configurations in Appendix~\ref{app:pairs}.  \emph{Real-robot evidence.}  The SO-ARM101 deployment covers five sorting tasks on one inexpensive 6-DOF arm ($50$ trials total) and validates feasibility without architectural modification, but it does not establish generalisation to other task families or to bimanual, mobile, or high-precision platforms.  \emph{Compatibility is an interpretation, not a demonstrated cause.}  The attention and weight-space analyses of Appendix~\ref{app:repalign} are consistent with compatibility transfer, but weight-space CKA is not activation CKA and cannot alone establish preserved representations or grounding. We manipulate compatibility only through initialization, so the causal mechanism remains open. Finally, the KV-share interface is held fixed across all transformer-block substrates, so the rules may not survive a different VLM--policy connector. Appendix~\ref{app:limitations} expands each point with concrete next steps.

\bibliography{example}

\newpage
\appendix

\section{Implementation and Training Details}
\label{app:implementation}

\paragraph{Substrate implementations.} The eight named substrates of Section~\ref{sec:exploration} are built as follows.
\begin{itemize}[leftmargin=1.2em]\setlength\itemsep{2pt}
\item \textbf{\textsc{Fast}} (autoregressive trunk): decodes action chunks as language tokens through Qwen2.5 itself, using the FAST frequency-domain action tokenizer~\citep{pertsch2025fast}.  Inference is standard autoregressive decoding through the language backbone; the substrate has no separate stack and \textsc{VLM-init} therefore does not apply.
\item \textbf{\textsc{OFT}} (MLP regression on VLM hidden states): reads VLM hidden states at the action positions through an MLP-ResNet head and predicts continuous action chunks with an $L_1$ loss.  The MLP has $10$M / $30$M / $80$M parameters at sizes $s$ / $m$ / $l$.  The architecture is not compatible with copying a transformer layer, so \textsc{VLM-init} does not apply.
\item \textbf{Transformer-block stack with KV-share.}  The four substrates \textsc{OFT-TF}, \textsc{EffVLA}, $\pi_0$, and $\pi_0$-\textsc{vlminit} all stack transformer blocks on top of the VLM and read the language backbone's KV cache directly via a KV-share interface, so the action stack does not duplicate the VLM forward pass.  The block hidden dimension is tied to the language-backbone hidden dimension; per-cell depths are tabulated in Table~\ref{tab:depth-width}.  Only $\pi_0$ uses the DiT variant~\citep{peebles2023scalable}, whose block carries adaptive layer-norm conditioning on the denoising timestep; \textsc{OFT-TF}, \textsc{EffVLA} and $\pi_0$-\textsc{vlminit} all use the plain block.  Within the $L_1$ pair this means the random-initialization control and \textsc{VLM-init} share an identical architecture and only the weights differ.
\item \textbf{\textsc{VLM-init}.} For \textsc{EffVLA} and $\pi_0$-\textsc{vlminit} we copy the parameter tensors of the last $D$ transformer layers of Qwen2.5-3B into the $D$ action-head layers.  Architectural compatibility (matching hidden size, head count, and attention pattern) is enforced by construction.
\item \textbf{Flow matching ($\pi_0$, $\pi_0$-\textsc{vlminit}).} The transformer stack is trained on a noisy action trajectory with the rectified-flow objective~\citep{lipman2022flow,liu2022flow} and integrated for four denoising passes at inference time, each applying a fixed-step update with the same parameters.  Two caveats apply to the flow-matching contrasts. First, $\pi_0$ uses the DiT block and $\pi_0$-\textsc{vlminit} the plain block, because the copied backbone layers have no adaptive layer-norm parameters to fill.  The two therefore differ in block type as well as in initialization, so unlike the $L_1$ pair this contrast is not initialization-only, and the mild $-2.2$-point effect of \textsc{VLM-init} under flow matching may reflect the change of block rather than a property of flow matching itself.  Second, we hold the schedule fixed at four passes and do not sweep the number of denoising steps.
\item \textbf{Four-pass $L_1$ (\textsc{OFT-TF-r4}, \textsc{OFT-TF-vlminit-r4}).} The same transformer stack is invoked four times at inference, with the output of pass $i$ feeding the input of pass $i\!+\!1$; training back-propagates through the four passes.  No noise schedule is used; the repeat is purely a compute match to the four-pass flow-matching schedule, so the loss-versus-passes contrast in Section~\ref{sec:exploration} is well-defined.
\end{itemize}

\paragraph{Two-stage training pipeline.} Every cell follows the same pipeline.  Stage~A (action-head pretraining) trains on DROID~\citep{khazatsky2024droid} with the public SigLIP2 and Qwen2.5 backbone checkpoints loaded but frozen, for the action-head architecture under study.  Stage~B (full-VLA fine-tuning) trains on LIBERO~\citep{liu2023libero} for $100$k optimizer steps with every module unfrozen.  Evaluation runs LIBERO-Plus~\citep{fei2025libero}: four task suites $\times$ seven perturbation axes, ${\sim}1{,}000$ rollouts per cell.  Optimizer, learning rate, batch size, data augmentation, and chunk size ($8$) are held identical across cells.

\paragraph{\textsc{EffVLA} specifics.} \textsc{EffVLA} is the \textsc{OFT-TF-vlminit} substrate at $V\!=\!L\!=\!A\!=\!l$: a $4$-layer transformer-block stack initialized from the last $4$ transformer layers of Qwen2.5-3B, on a SigLIP2-So400m vision encoder at $256$-pixel input.  The head adds $\sim\!0.35$B parameters, for a full model of $\sim\!3.75$B.  Training follows the same Stage~A$+$B pipeline.  For the SO-ARM101 real-robot deployment (Section~\ref{sec:efficientvla}) only the data pipeline changes: Stage~A uses the SO-ARM101 open-source community corpus in place of DROID, and Stage~B fine-tunes on teleoperated demonstrations of the five deployment tasks, merged into a single training set so that one model serves all five.

\section{Latency Measurement Protocol}
\label{app:latency-protocol}

\paragraph{Hardware and runtime.} All wall-clock latency measurements use an NVIDIA RTX~5090 in bf16 with batch size $1$ and action chunk size $8$.  Each model is compiled with \texttt{torch.compile} before timing; no further kernel fusion, custom CUDA kernels, or quantization is applied. KV-share is implemented by reusing the language backbone's KV cache via standard attention masking, so the transformer-block action stack adds neither a duplicate prefill nor a duplicate KV computation to the language forward pass.

\paragraph{Timing.} After compilation, each cell is timed over $100$ runs and we report the mean; the per-cell standard deviation is below $1$~ms in every measured cell. Differences below $1$~ms are treated as below the measurement-noise threshold throughout the paper.

\paragraph{What is and is not included.} The reported latency covers the full forward pass from a pre-tokenized image and instruction to the produced action chunk: vision encoding, language backbone forward, action-head forward (with the four-pass schedule integrated where applicable), and the FAST action de-tokenization in the autoregressive case.  Image preprocessing, environment interaction, and tokenizer lookup are excluded because they are deployment-target-dependent rather than properties of the model.

\section{Complete Cell-Level Results}
\label{app:results}

This section reports every trained cell at $100$k fine-tuning steps with the pretrained SigLIP2$+$Qwen2.5 backbone and the standard two-stage DROID$+$LIBERO pipeline the main text uses throughout.  We trained $63$ cells on the module-scale factorial: the $V\!=\!L\!=\!A\!=\!l$ anchor for every substrate, the per-module $\{s, m, l\}$ sweeps for the three deployment-relevant substrates (\textsc{OFT}, \textsc{EffVLA}, $\pi_0$), the larger-than-$l$ scales $V\!=\!g$, $L\!=\!g$, $A\!=\!g$ used to establish saturation (Finding~3), the $4$-pass repeats (\textsc{OFT-TF-r4}, \textsc{OFT-TF-vlminit-r4}), and the \textsc{Fast} $V\!\times\!L$ sweep that has no separate $A$ axis.  Table~\ref{tab:full-grid} additionally lists $4$ input-resolution variants of \textsc{OFT} and $\pi_0$ at $V\!=\!l_3, l_5$ (corresponding to $384$- and $512$-px input) that we include as context but do not use for the scaling analyses.  The cells we did not run are off-diagonal small-scale combinations that the per-axis analyses do not depend on.

Three views of the same cells appear below. Table~\ref{tab:full-grid} reports the LIBERO-Plus 4-suite mean per cell, Table~\ref{tab:full-grid-axis} reports results on the seven LIBERO-Plus perturbation axes, and Table~\ref{tab:full-grid-libero} reports the standard LIBERO success rate on the four canonical task suites. Measured wall-clock latencies for the cells we directly profiled are collected in Appendix~\ref{app:scaling} (Table~\ref{tab:depth-width}; the $V$- and $A$-sweep latencies under \textsc{EffVLA} are also marked in Figure~\ref{fig:module-budget}).

\begin{small}
\setlength{\tabcolsep}{4pt}
\begin{longtable}{l c c c c c c c c}
\caption{Complete LIBERO-Plus per-suite results for every trained cell at $100$k fine-tuning steps with the pretrained SigLIP2$+$Qwen2.5 backbone (\textbf{WithoutPretrain}=No).
For each cell we report success rate (\%) on the four standard task suites (\texttt{spatial}, \texttt{object}, \texttt{goal}, \texttt{long}) and the 4-suite mean ``Avg.''.  Measured wall-clock latencies for the cells we directly profiled are in Table~\ref{tab:depth-width} (Appendix~\ref{app:scaling}).\label{tab:full-grid}}\\
\toprule
Substrate & $V$ & $L$ & $A$ & Spatial & Object & Goal & Long & Avg. \\
\midrule
\endfirsthead
\toprule
Substrate & $V$ & $L$ & $A$ & Spatial & Object & Goal & Long & Avg. \\
\midrule
\endhead
\midrule
\multicolumn{9}{r}{\emph{continued on next page}} \\
\endfoot
\bottomrule
\endlastfoot
\textsc{Fast} & s & s & --- & 33.1 & 30.6 & 11.7 & 20.2 & 23.9 \\
\textsc{Fast} & s & m & --- & 42.0 & 41.2 & 24.0 & 26.8 & 33.5 \\
\textsc{Fast} & s & l & --- & 61.8 & 66.4 & 47.1 & 47.9 & 55.8 \\
\textsc{Fast} & m & s & --- & 48.0 & 47.3 & 25.0 & 32.2 & 38.1 \\
\textsc{Fast} & m & m & --- & 48.8 & 50.1 & 30.6 & 35.4 & 41.2 \\
\textsc{Fast} & m & l & --- & 56.7 & 55.2 & 40.6 & 40.3 & 48.2 \\
\textsc{Fast} & l & s & --- & 46.0 & 51.0 & 24.7 & 34.6 & 39.1 \\
\textsc{Fast} & l & m & --- & 57.2 & 58.0 & 39.3 & 45.3 & 50.0 \\
\textsc{Fast} & l & l & --- & 66.4 & 66.0 & 48.2 & 50.2 & 57.7 \\
\midrule
\textsc{OFT} & s & s & s & 74.6 & 66.2 & 67.9 & 57.0 & 66.4 \\
\textsc{OFT} & s & s & m & 71.9 & 67.5 & 62.5 & 58.0 & 65.0 \\
\textsc{OFT} & s & s & l & 72.6 & 69.1 & 69.7 & 62.3 & 68.4 \\
\textsc{OFT} & s & l & l & 80.8 & 73.6 & 72.9 & 68.3 & 73.9 \\
\textsc{OFT} & m & m & s & 81.8 & 75.7 & 70.9 & 72.0 & 75.1 \\
\textsc{OFT} & m & m & m & 82.6 & 77.5 & 81.9 & 66.6 & 77.2 \\
\textsc{OFT} & m & m & l & 79.7 & 74.0 & 76.1 & 69.7 & 74.9 \\
\textsc{OFT} & m & l & l & 79.2 & 80.1 & 79.4 & 71.5 & 77.6 \\
\textsc{OFT} & l & s & l & 79.9 & 68.9 & 69.9 & 67.8 & 71.6 \\
\textsc{OFT} & l & m & l & 81.1 & 78.8 & 78.5 & 73.0 & 77.9 \\
\textsc{OFT} & l & l & s & 86.1 & 79.7 & 79.5 & 71.6 & 79.2 \\
\textsc{OFT} & l & l & m & 80.0 & 73.0 & 76.5 & 72.5 & 75.5 \\
\textsc{OFT} & l & l & l & 82.8 & 80.1 & 79.0 & 74.0 & 79.0 \\
\textsc{OFT} & l & g & l & 87.1 & 74.2 & 79.1 & 72.4 & 78.2 \\
\textsc{OFT} & $l_3$ & l & l & 78.5 & 78.4 & 76.1 & 72.4 & 76.4 \\
\textsc{OFT} & $l_5$ & l & l & 79.8 & 77.6 & 77.7 & 70.3 & 76.4 \\
\textsc{OFT} & g & l & l & 81.0 & 78.4 & 76.9 & 74.3 & 77.7 \\
\midrule
\textsc{OFT-TF} & s & s & s & 76.8 & 74.1 & 64.3 & 61.5 & 69.2 \\
\textsc{OFT-TF} & s & l & l & 73.8 & 73.3 & 71.2 & 58.4 & 69.2 \\
\textsc{OFT-TF} & m & m & m & 76.2 & 70.2 & 68.4 & 58.2 & 68.3 \\
\textsc{OFT-TF} & m & l & l & 75.4 & 72.6 & 69.7 & 60.1 & 69.5 \\
\textsc{OFT-TF} & l & s & l & 80.1 & 74.3 & 72.4 & 65.2 & 73.0 \\
\textsc{OFT-TF} & l & m & l & 75.1 & 70.5 & 70.7 & 63.4 & 69.9 \\
\textsc{OFT-TF} & l & l & s & 76.6 & 73.5 & 64.6 & 63.7 & 69.6 \\
\textsc{OFT-TF} & l & l & m & 74.1 & 73.8 & 70.6 & 62.2 & 70.2 \\
\textsc{OFT-TF} & l & l & l & 78.8 & 71.4 & 76.0 & 64.4 & 72.7 \\
\textsc{OFT-TF} & l & g & l & 79.6 & 76.3 & 73.3 & 66.4 & 73.9 \\
\midrule
\textsc{OFT-TF-r4} & l & l & l & 81.8 & 75.4 & 71.7 & 70.3 & 74.8 \\
\midrule
\textsc{EffVLA} & s & l & l & 80.7 & 80.6 & 75.5 & 68.8 & 76.4 \\
\textsc{EffVLA} & m & l & l & 82.2 & 78.2 & 77.1 & 69.6 & 76.8 \\
\textsc{EffVLA} & l & s & l & 86.5 & 83.2 & 73.0 & 69.5 & 78.1 \\
\textsc{EffVLA} & l & m & l & 83.8 & 80.9 & 74.7 & 67.8 & 76.8 \\
\textsc{EffVLA} & l & l & s & 79.4 & 74.5 & 63.8 & 63.9 & 70.4 \\
\textsc{EffVLA} & l & l & m & 83.7 & 83.1 & 75.1 & 69.9 & 78.0 \\
\textsc{EffVLA} & l & l & l & 84.4 & 84.5 & 75.5 & 74.7 & 79.8 \\
\textsc{EffVLA} & l & l & g & 87.0 & 78.4 & 74.7 & 73.0 & 78.3 \\
\textsc{EffVLA} & l & g & l & 85.5 & 80.7 & 75.2 & 69.9 & 77.8 \\
\textsc{EffVLA} & g & l & l & 83.8 & 81.8 & 75.9 & 74.0 & 78.9 \\
\midrule
\textsc{OFT-TF-vlminit-r4} & l & l & l & 84.4 & 83.7 & 76.9 & 73.3 & 79.6 \\
\midrule
$\pi_0$ & s & s & s & 74.0 & 69.7 & 65.4 & 60.1 & 67.3 \\
$\pi_0$ & s & s & m & 75.6 & 64.3 & 61.4 & 58.4 & 64.9 \\
$\pi_0$ & s & s & l & 76.3 & 73.7 & 59.9 & 54.6 & 66.1 \\
$\pi_0$ & s & l & l & 78.1 & 74.6 & 72.1 & 68.1 & 73.2 \\
$\pi_0$ & m & m & s & 79.0 & 74.8 & 74.1 & 67.3 & 73.8 \\
$\pi_0$ & m & m & m & 84.2 & 76.0 & 73.6 & 67.3 & 75.3 \\
$\pi_0$ & m & m & l & 79.5 & 76.9 & 74.9 & 66.3 & 74.4 \\
$\pi_0$ & m & l & l & 81.4 & 77.0 & 76.3 & 67.8 & 75.6 \\
$\pi_0$ & l & s & l & 78.5 & 71.8 & 68.7 & 64.4 & 70.9 \\
$\pi_0$ & l & m & l & 80.1 & 70.7 & 72.9 & 67.5 & 72.8 \\
$\pi_0$ & l & l & s & 79.6 & 73.0 & 71.1 & 66.9 & 72.7 \\
$\pi_0$ & l & l & m & 78.5 & 73.0 & 73.6 & 65.3 & 72.6 \\
$\pi_0$ & l & l & l & 83.6 & 83.7 & 72.6 & 69.7 & 77.4 \\
$\pi_0$ & l & l & g & 80.6 & 76.3 & 76.9 & 70.5 & 76.1 \\
$\pi_0$ & l & g & l & 84.3 & 78.5 & 77.6 & 69.3 & 77.4 \\
$\pi_0$ & $l_3$ & l & l & 79.6 & 79.9 & 75.8 & 74.5 & 77.5 \\
$\pi_0$ & $l_5$ & l & l & 79.0 & 75.5 & 75.6 & 72.4 & 75.6 \\
$\pi_0$ & g & l & l & 83.8 & 79.2 & 76.6 & 70.9 & 77.6 \\
\midrule
$\pi_0$-\textsc{vlminit} & l & l & l & 81.6 & 76.7 & 73.4 & 68.9 & 75.2 \\
\end{longtable}
\end{small}

\begin{small}
\setlength{\tabcolsep}{3pt}
\begin{longtable}{l c c c c c c c c c c}
\caption{Complete LIBERO-Plus per-perturbation-axis results for every trained cell at $100$k fine-tuning steps with the pretrained backbone.
The seven axes are background textures (BG), robot initial states (Init), camera viewpoints (Cam), language instructions (Lang), sensor noise (Noise), object layouts (Layout), and light conditions (Light).  No aggregate column is reported here: the seven perturbation axes contain different numbers of episodes, so any unweighted mean over them would not be the LIBERO-Plus score.  The benchmark score per cell is the four-suite mean given in Table~\ref{tab:full-grid}.\label{tab:full-grid-axis}}\\
\toprule
Substrate & $V$ & $L$ & $A$ & BG & Init & Cam & Lang & Noise & Layout & Light \\
\midrule
\endfirsthead
\toprule
Substrate & $V$ & $L$ & $A$ & BG & Init & Cam & Lang & Noise & Layout & Light \\
\midrule
\endhead
\midrule
\multicolumn{11}{r}{\emph{continued on next page}} \\
\endfoot
\bottomrule
\endlastfoot
\textsc{Fast} & s & s & --- & 54.4 & 0.1 & 0.0 & 37.9 & 0.0 & 41.2 & 53.3 \\
\textsc{Fast} & s & m & --- & 65.2 & 0.8 & 0.3 & 58.1 & 0.1 & 60.1 & 72.6 \\
\textsc{Fast} & s & l & --- & 80.6 & 34.7 & 31.8 & 67.1 & 39.5 & 70.6 & 84.3 \\
\textsc{Fast} & m & s & --- & 75.0 & 6.5 & 6.6 & 58.1 & 3.2 & 64.7 & 78.2 \\
\textsc{Fast} & m & m & --- & 79.9 & 8.7 & 5.9 & 61.1 & 8.9 & 67.8 & 82.9 \\
\textsc{Fast} & m & l & --- & 82.7 & 17.7 & 19.4 & 70.6 & 17.1 & 70.6 & 82.5 \\
\textsc{Fast} & l & s & --- & 73.3 & 14.1 & 5.6 & 58.5 & 5.9 & 62.7 & 77.1 \\
\textsc{Fast} & l & m & --- & 83.6 & 24.3 & 22.1 & 64.2 & 18.1 & 75.1 & 85.5 \\
\textsc{Fast} & l & l & --- & 87.8 & 33.3 & 27.9 & 75.6 & 35.9 & 75.7 & 88.7 \\
\midrule
\textsc{OFT} & s & s & s & 91.8 & 63.3 & 34.6 & 72.1 & 48.9 & 79.4 & 93.1 \\
\textsc{OFT} & s & s & m & 94.1 & 59.6 & 36.2 & 69.3 & 45.4 & 77.1 & 92.3 \\
\textsc{OFT} & s & s & l & 93.8 & 60.0 & 39.7 & 74.1 & 56.8 & 78.1 & 93.6 \\
\textsc{OFT} & s & l & l & 96.3 & 67.2 & 56.6 & 84.7 & 50.5 & 81.5 & 96.1 \\
\textsc{OFT} & m & m & s & 91.9 & 67.7 & 52.9 & 77.7 & 74.5 & 78.9 & 94.7 \\
\textsc{OFT} & m & m & m & 96.9 & 68.8 & 53.2 & 63.4 & 72.6 & 80.9 & 96.4 \\
\textsc{OFT} & m & m & l & 98.5 & 66.0 & 49.0 & 80.2 & 67.2 & 81.0 & 98.4 \\
\textsc{OFT} & m & l & l & 95.5 & 63.7 & 61.1 & 85.4 & 71.5 & 82.3 & 96.0 \\
\textsc{OFT} & l & s & l & 96.3 & 64.6 & 42.3 & 76.6 & 61.1 & 80.6 & 96.4 \\
\textsc{OFT} & l & m & l & 97.0 & 66.9 & 57.8 & 83.2 & 73.4 & 83.2 & 96.0 \\
\textsc{OFT} & l & l & s & 96.6 & 68.5 & 59.6 & 89.3 & 72.3 & 83.4 & 97.7 \\
\textsc{OFT} & l & l & m & 97.5 & 65.9 & 51.7 & 86.0 & 60.8 & 83.7 & 98.8 \\
\textsc{OFT} & l & l & l & 97.8 & 68.0 & 53.8 & 85.3 & 79.4 & 82.9 & 98.3 \\
\textsc{OFT} & l & g & l & 97.2 & 65.6 & 68.7 & 84.8 & 64.6 & 84.4 & 96.0 \\
\textsc{OFT} & $l_3$ & l & l & 96.7 & 65.3 & 45.8 & 90.1 & 68.0 & 84.3 & 98.6 \\
\textsc{OFT} & $l_5$ & l & l & 98.5 & 67.3 & 46.6 & 91.1 & 65.2 & 83.2 & 97.8 \\
\textsc{OFT} & g & l & l & 97.4 & 64.5 & 57.9 & 83.0 & 74.9 & 82.3 & 97.2 \\
\midrule
\textsc{OFT-TF} & s & s & s & 96.7 & 54.7 & 39.5 & 84.6 & 52.5 & 78.8 & 95.8 \\
\textsc{OFT-TF} & s & l & l & 94.7 & 50.7 & 48.8 & 87.0 & 48.4 & 78.2 & 93.9 \\
\textsc{OFT-TF} & m & m & m & 94.6 & 50.4 & 43.9 & 85.4 & 46.4 & 81.7 & 94.3 \\
\textsc{OFT-TF} & m & l & l & 96.6 & 54.4 & 47.3 & 84.9 & 44.9 & 80.8 & 95.8 \\
\textsc{OFT-TF} & l & s & l & 97.6 & 60.3 & 50.0 & 83.5 & 55.2 & 82.9 & 99.5 \\
\textsc{OFT-TF} & l & m & l & 97.8 & 53.6 & 41.4 & 89.3 & 46.9 & 82.1 & 97.9 \\
\textsc{OFT-TF} & l & l & s & 94.5 & 56.4 & 42.6 & 84.5 & 48.4 & 81.4 & 96.7 \\
\textsc{OFT-TF} & l & l & m & 95.9 & 58.7 & 48.6 & 81.7 & 45.1 & 81.6 & 98.3 \\
\textsc{OFT-TF} & l & l & l & 97.0 & 57.8 & 53.0 & 86.9 & 41.8 & 62.5 & 74.0 \\
\textsc{OFT-TF} & l & g & l & 99.1 & 55.5 & 60.4 & 86.8 & 52.5 & 81.6 & 98.7 \\
\midrule
\textsc{OFT-TF-r4} & l & l & l & 97.7 & 60.7 & 55.8 & 88.5 & 56.4 & 81.8 & 99.0 \\
\midrule
\textsc{EffVLA} & s & l & l & 96.7 & 62.4 & 67.3 & 88.0 & 55.8 & 81.4 & 98.1 \\
\textsc{EffVLA} & m & l & l & 97.4 & 64.3 & 63.4 & 84.9 & 62.5 & 81.4 & 98.5 \\
\textsc{EffVLA} & l & s & l & 98.2 & 65.3 & 62.8 & 84.2 & 70.7 & 81.2 & 97.4 \\
\textsc{EffVLA} & l & m & l & 97.1 & 61.3 & 64.8 & 83.5 & 65.1 & 83.5 & 96.0 \\
\textsc{EffVLA} & l & l & s & 98.1 & 49.5 & 49.0 & 85.5 & 50.4 & 81.0 & 98.2 \\
\textsc{EffVLA} & l & l & m & 98.4 & 67.4 & 65.1 & 87.6 & 60.3 & 83.1 & 98.6 \\
\textsc{EffVLA} & l & l & l & 96.7 & 69.8 & 68.3 & 87.5 & 66.1 & 84.8 & 97.4 \\
\textsc{EffVLA} & l & l & g & 96.5 & 65.4 & 65.9 & 85.7 & 65.7 & 84.0 & 98.2 \\
\textsc{EffVLA} & l & g & l & 97.1 & 58.9 & 64.6 & 89.3 & 66.8 & 83.6 & 98.1 \\
\textsc{EffVLA} & g & l & l & 97.1 & 70.9 & 61.7 & 87.5 & 67.3 & 84.3 & 95.6 \\
\midrule
\textsc{OFT-TF-vlminit-r4} & l & l & l & 98.0 & 67.2 & 64.9 & 89.1 & 68.2 & 85.1 & 97.4 \\
\midrule
$\pi_0$ & s & s & s & 93.3 & 55.3 & 43.6 & 77.9 & 44.2 & 79.4 & 96.4 \\
$\pi_0$ & s & s & m & 92.5 & 52.4 & 40.4 & 79.3 & 31.1 & 60.7 & 70.2 \\
$\pi_0$ & s & s & l & 92.1 & 48.4 & 48.0 & 74.4 & 46.9 & 78.6 & 92.6 \\
$\pi_0$ & s & l & l & 95.8 & 64.2 & 54.3 & 83.6 & 52.0 & 82.2 & 96.2 \\
$\pi_0$ & m & m & s & 95.9 & 61.9 & 45.7 & 86.2 & 63.4 & 82.4 & 97.1 \\
$\pi_0$ & m & m & m & 98.5 & 65.6 & 53.5 & 85.9 & 59.7 & 82.1 & 97.6 \\
$\pi_0$ & m & m & l & 97.9 & 64.7 & 53.4 & 88.2 & 54.2 & 81.8 & 96.7 \\
$\pi_0$ & m & l & l & 98.7 & 68.3 & 58.1 & 84.9 & 56.4 & 81.5 & 97.6 \\
$\pi_0$ & l & s & l & 96.4 & 59.1 & 46.7 & 82.9 & 48.2 & 82.7 & 98.8 \\
$\pi_0$ & l & m & l & 97.6 & 61.5 & 49.5 & 85.7 & 51.0 & 84.0 & 97.9 \\
$\pi_0$ & l & l & s & 97.0 & 54.4 & 49.2 & 90.4 & 53.5 & 83.7 & 98.0 \\
$\pi_0$ & l & l & m & 97.7 & 62.8 & 48.9 & 81.6 & 52.7 & 83.8 & 98.6 \\
$\pi_0$ & l & l & l & 97.1 & 67.4 & 62.9 & 84.6 & 64.7 & 82.1 & 96.4 \\
$\pi_0$ & l & l & g & 98.7 & 67.4 & 53.2 & 87.9 & 57.5 & 85.1 & 98.8 \\
$\pi_0$ & l & g & l & 97.6 & 66.5 & 66.0 & 85.7 & 59.1 & 84.5 & 96.9 \\
$\pi_0$ & $l_3$ & l & l & 98.1 & 66.9 & 59.0 & 87.0 & 57.7 & 83.9 & 97.1 \\
$\pi_0$ & $l_5$ & l & l & 96.8 & 68.2 & 50.9 & 90.1 & 54.0 & 86.0 & 99.3 \\
$\pi_0$ & g & l & l & 98.0 & 65.5 & 58.4 & 84.7 & 67.2 & 85.5 & 98.6 \\
\midrule
$\pi_0$-\textsc{vlminit} & l & l & l & 96.8 & 66.8 & 54.7 & 85.9 & 53.3 & 87.0 & 97.7 \\
\end{longtable}
\end{small}

\begin{small}
\setlength{\tabcolsep}{4pt}
\begin{longtable}{l c c c c c c c c}
\caption{Standard LIBERO success rates (\%) for every trained cell at $100$k fine-tuning steps with the pretrained backbone.  Columns are the four canonical task suites and their unweighted mean.\label{tab:full-grid-libero}}\\
\toprule
Substrate & $V$ & $L$ & $A$ & Spatial & Object & Goal & Long & Avg. \\
\midrule
\endfirsthead
\toprule
Substrate & $V$ & $L$ & $A$ & Spatial & Object & Goal & Long & Avg. \\
\midrule
\endhead
\midrule
\multicolumn{9}{r}{\emph{continued on next page}} \\
\endfoot
\bottomrule
\endlastfoot
\textsc{Fast} & s & s & --- & 86.0 & 95.0 & 81.6 & 82.2 & 86.2 \\
\textsc{Fast} & s & m & --- & 89.4 & 96.8 & 88.2 & 81.0 & 88.9 \\
\textsc{Fast} & s & l & --- & 89.8 & 98.2 & 90.4 & 84.0 & 90.6 \\
\textsc{Fast} & m & s & --- & 91.6 & 98.2 & 88.8 & 89.0 & 91.9 \\
\textsc{Fast} & m & m & --- & 92.6 & 96.6 & 90.6 & 83.2 & 90.8 \\
\textsc{Fast} & m & l & --- & 92.8 & 96.0 & 88.2 & 88.6 & 91.4 \\
\textsc{Fast} & l & s & --- & 92.8 & 97.4 & 87.0 & 86.4 & 90.9 \\
\textsc{Fast} & l & m & --- & 89.6 & 98.4 & 92.2 & 88.8 & 92.3 \\
\textsc{Fast} & l & l & --- & 91.2 & 98.2 & 92.0 & 90.2 & 92.9 \\
\midrule
\textsc{OFT} & s & s & s & 97.4 & 99.8 & 98.4 & 92.2 & 97.0 \\
\textsc{OFT} & s & s & m & 97.6 & 99.0 & 97.6 & 92.0 & 96.6 \\
\textsc{OFT} & s & s & l & 96.4 & 99.6 & 97.8 & 92.6 & 96.6 \\
\textsc{OFT} & s & l & l & 98.8 & 99.2 & 99.0 & 92.4 & 97.4 \\
\textsc{OFT} & m & m & s & 97.8 & 99.6 & 98.2 & 93.2 & 97.2 \\
\textsc{OFT} & m & m & m & 98.8 & 98.8 & 99.2 & 93.2 & 97.5 \\
\textsc{OFT} & m & m & l & 98.6 & 99.6 & 98.6 & 96.0 & 98.2 \\
\textsc{OFT} & m & l & l & 98.8 & 98.6 & 98.8 & 95.6 & 98.0 \\
\textsc{OFT} & l & s & l & 97.8 & 98.8 & 97.6 & 95.0 & 97.3 \\
\textsc{OFT} & l & m & l & 99.2 & 99.0 & 98.2 & 97.6 & 98.5 \\
\textsc{OFT} & l & l & s & 99.2 & 99.2 & 97.8 & 92.6 & 97.2 \\
\textsc{OFT} & l & l & m & 98.8 & 98.8 & 98.8 & 95.0 & 97.9 \\
\textsc{OFT} & l & l & l & 98.8 & 99.0 & 98.0 & 93.6 & 97.4 \\
\textsc{OFT} & $l_3$ & l & l & 98.2 & 99.0 & 99.0 & 95.0 & 97.8 \\
\textsc{OFT} & $l_5$ & l & l & 97.2 & 99.0 & 97.8 & 95.2 & 97.3 \\
\midrule
\textsc{OFT-TF} & s & s & s & 98.4 & 99.0 & 98.2 & 95.4 & 97.8 \\
\textsc{OFT-TF} & s & l & l & 97.4 & 98.2 & 98.6 & 94.2 & 97.1 \\
\textsc{OFT-TF} & m & m & m & 99.4 & 98.8 & 99.0 & 95.0 & 98.1 \\
\textsc{OFT-TF} & m & l & l & 97.6 & 98.6 & 99.0 & 95.6 & 97.7 \\
\textsc{OFT-TF} & l & s & l & 98.8 & 99.8 & 99.2 & 94.6 & 98.1 \\
\textsc{OFT-TF} & l & m & l & 99.0 & 99.6 & 98.6 & 95.2 & 98.1 \\
\textsc{OFT-TF} & l & l & s & 99.2 & 99.4 & 99.2 & 97.2 & 98.8 \\
\textsc{OFT-TF} & l & l & m & 98.8 & 99.6 & 99.0 & 96.4 & 98.5 \\
\textsc{OFT-TF} & l & l & l & 98.2 & 98.4 & 99.4 & 96.6 & 98.2 \\
\midrule
\textsc{OFT-TF-r4} & l & l & l & 99.2 & 99.4 & 98.6 & 97.2 & 98.6 \\
\midrule
\textsc{EffVLA} & l & l & l & 98.6 & 99.2 & 99.0 & 96.0 & 98.2 \\
\midrule
\textsc{OFT-TF-vlminit-r4} & l & l & l & 98.4 & 99.2 & 99.2 & 95.2 & 98.0 \\
\midrule
$\pi_0$ & s & s & s & 98.6 & 99.0 & 99.0 & 96.6 & 98.3 \\
$\pi_0$ & s & s & m & 98.0 & 99.8 & 99.0 & 95.4 & 98.1 \\
$\pi_0$ & s & s & l & 97.2 & 99.4 & 98.6 & 93.6 & 97.2 \\
$\pi_0$ & s & l & l & 97.4 & 99.8 & 97.8 & 96.0 & 97.8 \\
$\pi_0$ & m & m & s & 99.2 & 99.4 & 98.4 & 96.4 & 98.4 \\
$\pi_0$ & m & m & m & 99.2 & 99.4 & 99.4 & 97.4 & 98.9 \\
$\pi_0$ & m & m & l & 97.6 & 99.4 & 98.8 & 94.8 & 97.7 \\
$\pi_0$ & m & l & l & 98.2 & 99.6 & 99.2 & 95.0 & 98.0 \\
$\pi_0$ & l & s & l & 99.0 & 98.8 & 98.8 & 95.6 & 98.1 \\
$\pi_0$ & l & m & l & 98.8 & 99.2 & 98.2 & 96.6 & 98.2 \\
$\pi_0$ & l & l & s & 99.4 & 99.8 & 99.2 & 96.4 & 98.7 \\
$\pi_0$ & l & l & m & 98.0 & 99.2 & 98.4 & 95.8 & 97.9 \\
$\pi_0$ & l & l & l & 98.6 & 99.6 & 98.2 & 95.2 & 97.9 \\
$\pi_0$ & l & l & g & 99.4 & 99.6 & 99.0 & 96.6 & 98.7 \\
$\pi_0$ & $l_3$ & l & l & 98.6 & 99.4 & 98.2 & 97.0 & 98.3 \\
$\pi_0$ & $l_5$ & l & l & 99.2 & 99.4 & 99.2 & 94.4 & 98.1 \\
\midrule
$\pi_0$-\textsc{vlminit} & l & l & l & 99.0 & 99.0 & 99.2 & 95.4 & 98.2 \\
\end{longtable}
\end{small}

\section{Module Scaling Analysis}
\label{app:scaling}

This section breaks the aggregate Figure~\ref{fig:module-scaling} sweep down two ways and characterises the architectural caveat that the $\rho$ values in Section~\ref{sec:scaling} carry.  In both breakdowns the three substrates and the colour scheme of Figure~\ref{fig:module-scaling} are kept; reading off any panel reproduces the substrate ordering visible in the aggregate.

\begin{figure}[t]
\centering
\includegraphics[width=\linewidth]{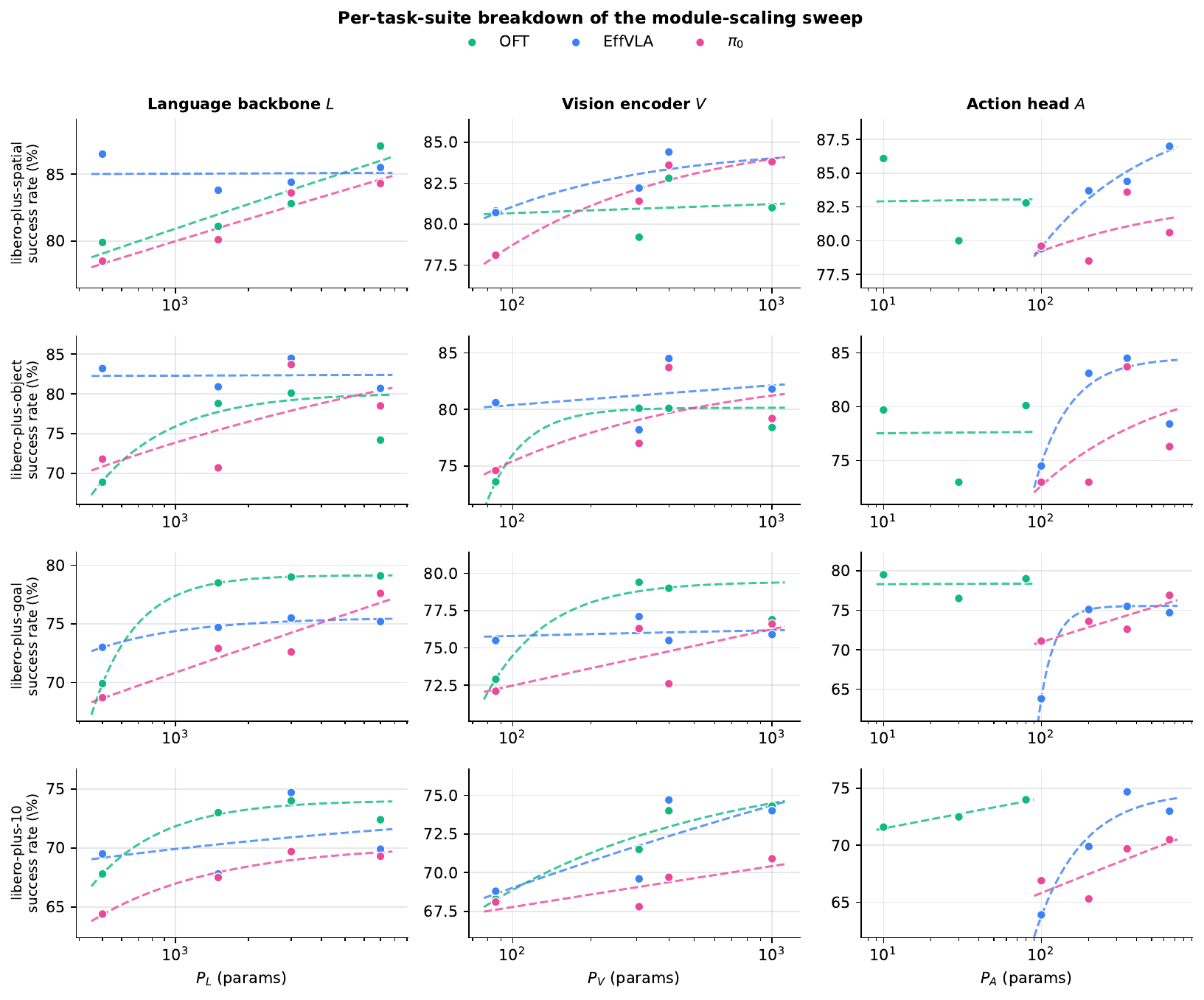}
\caption{Per-task-suite breakdown of the module-scaling sweep.  Rows are the four LIBERO-Plus task suites (\texttt{spatial}, \texttt{object}, \texttt{goal}, \texttt{long-10}); columns are the three module-scaling axes ($L$, $V$, $A$).  Substrate curves match Figure~\ref{fig:module-scaling}: \textsc{OFT} (teal), \textsc{OFT-TF-vlminit} = \textsc{EffVLA} (blue), $\pi_0$ (pink); dashed lines are saturating fits through the measured scales.}
\label{fig:scaling-per-suite}
\end{figure}

\paragraph{Per-task-suite breakdown.}  Figure~\ref{fig:scaling-per-suite} marginalises the same module-scaling sweep over the four LIBERO-Plus task suites separately.  The aggregate ordering of Figure~\ref{fig:module-scaling} holds within every suite: the action head ($A$) axis separates \textsc{EffVLA} from \textsc{OFT} and $\pi_0$ on every suite, with the steepest substrate-specific rises sitting on the longer-horizon \texttt{long}-$10$ suite where harder tasks leave the most room for the head to help.  No suite reverses the qualitative picture: action-head scaling under \textsc{VLM-init} is the dominant per-suite trend, and the language and vision sweeps remain relatively flat once the head is fixed.

\begin{figure}[t]
\centering
\includegraphics[width=\linewidth,height=0.82\textheight,keepaspectratio]{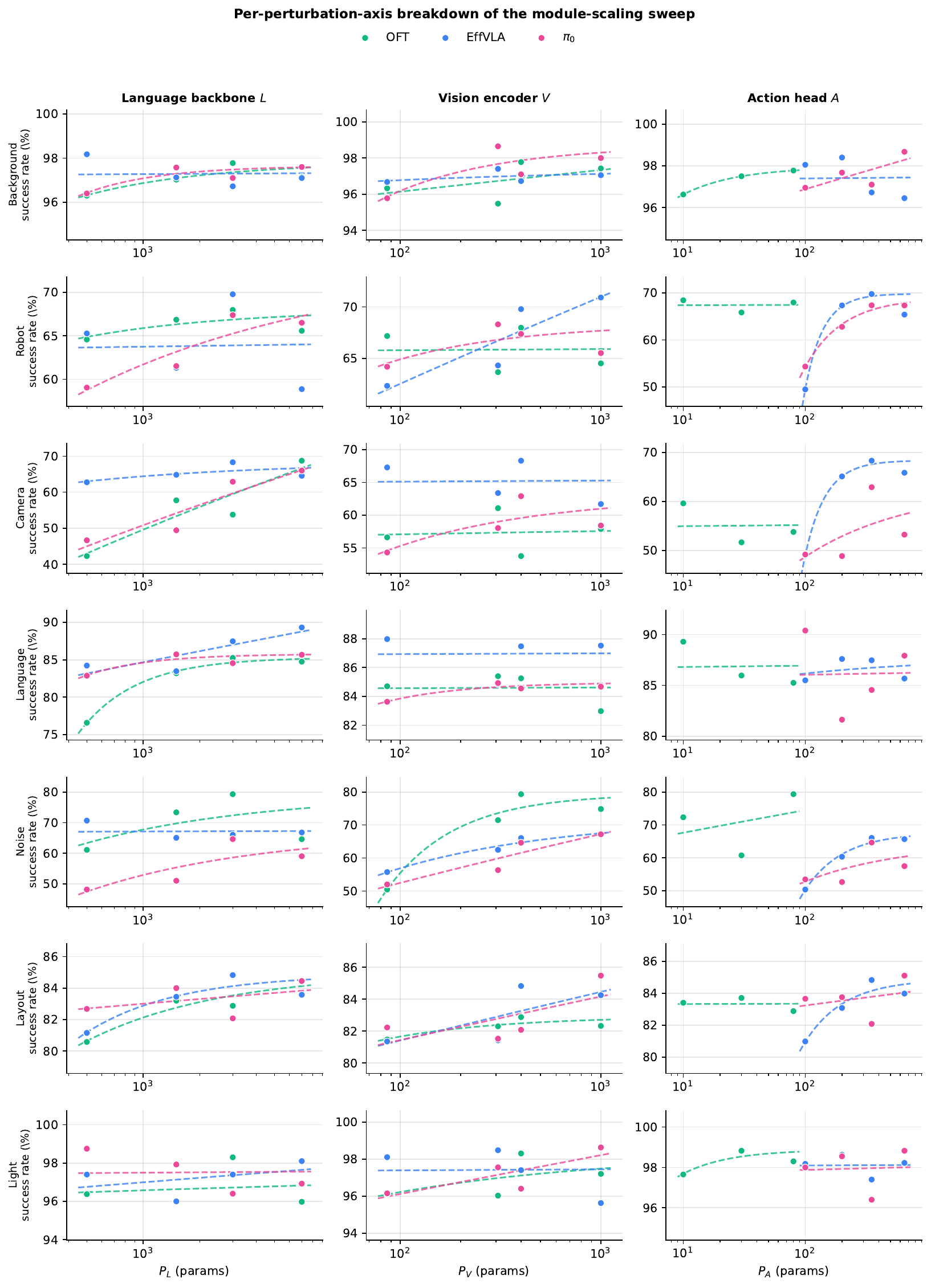}
\caption{Per-perturbation-axis breakdown of the module-scaling sweep.  Rows are the seven LIBERO-Plus perturbation axes (background textures, robot initial states, camera viewpoints, language instructions, sensor noise, object layouts, lighting); columns are the three module-scaling axes ($L$, $V$, $A$).  Substrate curves match Figure~\ref{fig:module-scaling}.  The per-axis sweep log is most complete for \textsc{OFT} and $\pi_0$; \textsc{EffVLA} carries fewer per-axis sweep points than its aggregate sweep.}
\label{fig:scaling-per-axis}
\end{figure}

\paragraph{Per-perturbation-axis breakdown.}  Figure~\ref{fig:scaling-per-axis} reorganises the same data along the seven LIBERO-Plus perturbation axes, and shows that the gains module scaling buys are concentrated rather than uniform.  On the visually-saturated axes the pretrained backbones already handle (background textures, object layout, lighting), every module's curve is flat to within a few points; almost all of the headroom sits on the harder axes (camera viewpoint, sensor noise, robot initial state).  The distribution of gains across modules is itself informative: scaling $L$ helps camera viewpoint and language instructions most, scaling $V$ helps sensor noise most under the regression-head substrates, and scaling $A$ under \textsc{EffVLA} is the strongest single lever on robot initial state and camera viewpoint.  A practitioner targeting a specific perturbation weakness can read off the relevant panel rather than treat module size as a uniform efficiency lever.

\paragraph{Depth--width coupling under KV-share.}  On the four KV-share substrates the action stack's hidden dimension is tied to the language backbone's hidden dimension $h_L$, so hitting a comparable $A\!=\!l$ parameter target across $L$ scales forces the head depth to scale as $D \propto h_L^{-2}$.  Table~\ref{tab:depth-width} lists the configured depths and the resulting latencies and success rates for every $L$ scale at $V\!=\!A\!=\!l$.  \textsc{EffVLA}'s single-pass schedule keeps the $L$--latency relationship monotone; $\pi_0$'s four-pass schedule multiplies the depth penalty $4\times$ and inverts it, so $L\!=\!s$ becomes slower than $L\!=\!l$ overall.  The $\rho$ value for the $L$ axis under $\pi_0$ that Section~\ref{sec:scaling} would otherwise report therefore mixes the language-backbone effect with this architectural depth--width coupling, and we do not summarise it with a single slope.

\begin{table}[htbp]
\centering\small
\setlength{\tabcolsep}{6pt}
\caption{Depth--width coupling under KV-share transformer-block action heads.  At $V\!=\!A\!=\!l$ the head's hidden dimension is tied to the language backbone's hidden dimension $h_L$, so hitting a comparable $A\!=\!l$ parameter target across $L$ scales requires the head depth to scale as $D \propto h_L^{-2}$.  The latency consequence is amplified by the per-action-chunk pass count: \textsc{EffVLA}'s single-pass schedule keeps the $L$--latency relationship monotone, while $\pi_0$'s four-pass schedule makes a smaller $L$ \emph{slower} overall.}
\label{tab:depth-width}
\begin{tabular}{l c c c c c}
\toprule
Substrate & $L$ scale & $h_L$ & Depth $D$ & Latency (ms) & Succ.\ (\%) \\
\midrule
\textsc{EffVLA} & s & 896 & 24 & 27.8 & 78.1 \\
 & m & 1536 & 10 & 30.6 & 76.8 \\
 & l & 2048 & 4 & 39.2 & 79.8 \\
\midrule
$\pi_0$ & s & 896 & 24 & 76.5 & 70.9 \\
 & m & 1536 & 10 & 53.8 & 72.8 \\
 & l & 2048 & 6 & 53.7 & 77.4 \\
\bottomrule
\end{tabular}
\end{table}

\section{Effect of Pretraining}
\label{app:pretrain}

This is the only table in the appendix that reports cells trained without DROID action-head pretraining (Stage~A); every other result in the paper uses the standard two-stage pipeline with Stage~A.  Table~\ref{tab:pretrain} reports the LIBERO-Plus 4-suite mean for every cell that we ran both with and without DROID action-head pretraining (Stage~A).  Two patterns hold.  First, the $L_1$ family (\textsc{OFT}) is largely robust to dropping Stage~A: typical drops are $1$ to $6$ points, with the smallest scales showing the smallest gap (and one case of mild reversal).  Second, the flow-matching substrate ($\pi_0$) consistently loses $15$ to $22$ points across scales when Stage~A is removed.  This contrast is the basis of the Section~\ref{sec:efficientvla} statement that \textsc{EffVLA} inherits the $L_1$ family's pretraining robustness, which in turn is what makes the small Stage~B sufficient when the recipe is retargeted to a new embodiment (Section~\ref{sec:efficientvla}, real-robot deployment on SO-ARM101).

\begin{table}[htbp]
\centering\small
\setlength{\tabcolsep}{6pt}
\caption{Effect of removing DROID action-head pretraining (Stage~A) on LIBERO-Plus 4-suite mean success rate (\%).  ``With'' is the standard two-stage pipeline used throughout the paper (\textbf{WithoutPretrain}=No); ``Without'' skips Stage~A and fine-tunes the action head directly on LIBERO.  Backbones are loaded from public checkpoints in both columns.  This is the one table in the appendix that uses the without-pretrain rows.}
\label{tab:pretrain}
\begin{tabular}{l c c c c c c}
\toprule
Substrate & $V$ & $L$ & $A$ & With Stage A & Without Stage A & $\Delta$ (pts) \\
\midrule
\textsc{OFT} & s & s & s & 66.4 & 67.9 & $-1.50$ \\
\textsc{OFT} & m & m & m & 77.2 & 70.8 & $+6.40$ \\
\textsc{OFT} & l & l & m & 75.5 & --- & --- \\
\textsc{OFT} & l & l & l & 79.0 & 76.9 & $+2.10$ \\
\midrule
$\pi_0$ & s & s & s & 67.3 & 53.6 & $+13.70$ \\
$\pi_0$ & m & m & m & 75.3 & 56.7 & $+18.60$ \\
$\pi_0$ & l & l & l & 77.4 & 55.8 & $+21.60$ \\
\bottomrule
\end{tabular}
\end{table}

\section{Cost of Deviating from the Recipe}
\label{app:deviations}

Table~\ref{tab:deviations} lists the cost of deviating from \textsc{EffVLA} along each single axis, measured against the recipe's anchor ($79.8\%$ on LIBERO-Plus, $39.2$~ms on RTX~5090).  Every entry is one cell of the Section~\ref{sec:exploration} factorial, so no additional runs are required.  The rows are grouped into two blocks.  \emph{Substrate} captures the head-design axes the main text Findings~1 and~2 are about: dropping \textsc{VLM-init}, replacing $L_1$ by flow matching, replacing the transformer stack by an MLP, and running four passes instead of one.  \emph{Module scale} captures the per-axis cost of shrinking $A$, $L$, or $V$ from $l$ while leaving the other two at $l$.

\begin{table}[htbp]
\centering\small
\setlength{\tabcolsep}{5pt}
\caption{Cost of deviating from \textsc{EffVLA} along each design axis, taken directly from the Section~\ref{sec:exploration} factorial with no additional runs.  The anchor is \textsc{EffVLA} at $V\!=\!L\!=\!A\!=\!l$ ($79.78\%$, $39.24$~ms).  Each row reports the LIBERO-Plus 4-suite mean and measured latency of the deviated cell, and its difference from the anchor.  Negative $\Delta$ succ.\ is a drop relative to \textsc{EffVLA}.}
\label{tab:deviations}
\begin{tabular}{@{}l c c c c@{}}
\toprule
Deviation & Succ.\ (\%) & $\Delta$ succ. & Lat.\ (ms) & $\Delta$ lat. \\
\midrule
\multicolumn{5}{l}{\emph{Substrate (head design).}} \\
Replace \textsc{VLM-init} with random init (\textsc{OFT-TF} at $l,l,l$) & 72.7 & $-7.10$ & 39.9 & $+0.61$ \\
Replace L1 with flow matching ($\pi_0$ at $l,l,l$) & 77.4 & $-2.40$ & 53.7 & $+14.48$ \\
Replace transformer head with MLP (\textsc{OFT} at $l,l,l$) & 79.0 & $-0.80$ & 40.1 & $+0.81$ \\
Run four passes instead of one (\textsc{OFT-TF-vlminit-r4} at $l,l,l$) & 79.6 & $-0.20$ & --- & --- \\
\midrule
\multicolumn{5}{l}{\emph{Module scale.}} \\
Shrink action head $A\!:\!l\!\to\!s$ (\textsc{EffVLA}) & 70.4 & $-9.40$ & 36.9 & $-2.36$ \\
Shrink action head $A\!:\!l\!\to\!m$ (\textsc{EffVLA}) & 78.0 & $-1.80$ & 38.4 & $-0.89$ \\
Shrink language backbone $L\!:\!l\!\to\!s$ (\textsc{EffVLA}) & 78.1 & $-1.70$ & 27.8 & $-11.44$ \\
Shrink vision encoder $V\!:\!l\!\to\!s$ (\textsc{EffVLA}) & 76.4 & $-3.40$ & 36.4 & $-2.80$ \\
\bottomrule
\end{tabular}
\end{table}

\section{VLM Backbone Capability}
\label{app:vlm}

For context only: Table~\ref{tab:vlm} reports general multimodal capability scores of the (SigLIP2, Qwen2.5) backbone pairs used in our sweep on six standard VLM benchmarks (MMBench-EN/CN, SEED, OCRBench, TextVQA, MMMU, AI2D, HallusionBench).  Higher is better.  These numbers are not used as a metric in any of the paper's claims; they are included so a reader can map our $\{V_{s,m,l}, L_{s,m,l}\}$ labels to their corresponding off-the-shelf VLM capability, and confirm that the larger backbones we sweep over are genuinely more capable on standard multimodal tasks.

\begin{table}[htbp]
\centering\small
\setlength{\tabcolsep}{4pt}
\caption{General multimodal capability of the SigLIP2$+$Qwen2.5 backbone pairs used in the paper, on eight standard VLM benchmarks (MMBench-EN/CN, SEED, OCRBench, TextVQA, MMMU, AI2D, HallusionBench).  Each row is one $(V, L)$ pair from our sweep.  ``Avg.'' is the unweighted mean across the eight benchmarks.  Higher is better.  These scores provide context for the module-scaling choices made in the main text; they are not used as a metric in any of the paper's claims.}
\label{tab:vlm}
\begin{tabular}{c c c c c c c c c c c}
\toprule
$V$ & $L$ & MMB-EN & MMB-CN & SEED & OCR & TextVQA & MMMU & AI2D & Hallu. & Avg. \\
\midrule
S & S & 55.6 & 52.2 & 62.1 & 31.0 & 27.2 & 31.4 & 49.0 & 28.3 & \textbf{42.1} \\
S & M & 67.8 & 65.9 & 68.1 & 34.1 & 30.3 & 39.1 & 65.6 & 31.1 & \textbf{50.2} \\
S & L & 72.3 & 69.9 & 69.6 & 34.0 & 32.0 & 39.0 & 70.2 & 31.3 & \textbf{52.3} \\
M & S & 58.3 & 53.4 & 63.7 & 37.5 & 33.7 & 34.0 & 49.3 & 28.3 & \textbf{44.8} \\
M & M & 72.3 & 68.9 & 70.3 & 41.1 & 36.4 & 39.0 & 67.4 & 31.6 & \textbf{53.4} \\
M & L & 74.4 & 73.5 & 71.1 & 42.5 & 38.0 & 39.3 & 71.6 & 32.4 & \textbf{55.4} \\
L & S & 59.7 & 54.2 & 64.9 & 42.9 & 36.6 & 33.1 & 55.8 & 29.3 & \textbf{47.1} \\
L & M & 73.2 & 69.4 & 70.4 & 44.9 & 41.0 & 39.5 & 69.1 & 34.6 & \textbf{55.3} \\
L & L & 76.0 & 74.0 & 72.0 & 44.3 & 41.5 & 39.1 & 72.8 & 33.8 & \textbf{56.7} \\
G & S & 61.9 & 56.8 & 65.6 & 42.8 & 36.5 & 33.7 & 55.8 & 30.2 & \textbf{47.9} \\
G & M & 73.7 & 69.9 & 71.3 & 45.5 & 41.4 & 38.5 & 56.0 & 31.8 & \textbf{53.5} \\
G & L & 78.0 & 74.6 & 72.6 & 47.9 & 42.0 & 42.1 & 73.3 & 33.5 & \textbf{58.0} \\
L(256) & L & 76.0 & 74.0 & 72.0 & 44.3 & 41.5 & 39.1 & 72.8 & 33.8 & \textbf{56.7} \\
L(384) & L & 77.7 & 77.0 & 73.4 & 56.8 & 55.6 & 40.8 & 71.3 & 38.7 & \textbf{61.4} \\
L(512) & L & 77.1 & 76.4 & 72.9 & 63.0 & 60.8 & 41.7 & 75.5 & 39.6 & \textbf{63.4} \\
\bottomrule
\end{tabular}
\end{table}

\section{Cross-Benchmark Replication on SimplerEnv}
\label{app:simplerenv}

We repeat the substrate comparison on a second benchmark and embodiment, SimplerEnv on a
WidowX arm~\citep{li2024evaluating}, with the action-head axes unchanged.  Four substrates,
$24$ trials per task over $4$ tasks, $96$ trials per model.

\begin{table}[htbp]
\centering
\caption{\textbf{SimplerEnv (Bridge, WidowX): the same substrate ordering, on a different
embodiment.}  Success rate (\%); $24$ trials per task, $96$ per model.  The initialization
contrast reproduces: $+7.3$ points here against $+7.1$ on LIBERO-Plus.}
\label{tab:simplerenv}
\begin{tabular}{lccccc}
\toprule
Substrate & Stack & Carrot & Spoon & Eggplant & Overall \\
\midrule
\textsc{OFT} (MLP)                      & 50.0 & 37.5 & 50.0 & \textbf{100.0} & 59.4 \\
Transformer block, random init          & \phantom{0}8.3 & \textbf{58.3} & \textbf{79.2} & 70.8 & 54.2 \\
$\pi_0$ (flow matching)                 & 43.6 & 36.6 & 67.9 & 93.8 & 60.5 \\
\textsc{EffVLA} (transformer, \textsc{VLM-init}) & 37.5 & 50.0 & 75.0 & 83.3 & \textbf{61.5} \\
\bottomrule
\end{tabular}
\end{table}

The randomly initialized head here is architecture-matched: the same Qwen2.5 decoder layers are
deep-copied and then re-drawn, so topology, parameter count and latency are identical to
\textsc{EffVLA} and only the weight values differ.  That contrast is $54.2$ against $61.5$,
$+7.3$ points, next to $+7.1$ on LIBERO-Plus, which places the gain in the weights rather than
in the head's architecture or capacity.  The three strongest substrates again fall within a
few points of each other ($61.5$, $60.5$, $59.4$) and \textsc{OFT} and $\pi_0$ swap order
between benchmarks, so we read them as tied here as well.  At $96$ trials per model the
standard error on a difference is ${\approx}7$ points, so this is a replication of the effect
size and the ordering, not an independent significance test.

\section{Repeated-Seed Check on the Anchor Pair}
\label{app:seeds}

The sweep trains one seed per cell, so we re-trained the $V\!=\!L\!=\!A\!=\!l$ anchor pair from
scratch under five seeds to measure run-to-run variation directly.  Seed~1 is the run reported
throughout the paper.

\begin{table}[htbp]
\centering
\caption{\textbf{Five training seeds for the close \textsc{EffVLA}--\textsc{OFT} comparison at $l/l/l$.} LIBERO-Plus mean success rate (\%). The repeated runs test whether the small single-run ordering is stable to training seed; Seed~1 is the run used elsewhere in the paper.}
\label{tab:seeds}
\begin{tabular}{cccc}
\toprule
Seed & \textsc{EffVLA} & \textsc{OFT} & $\Delta$ \\
\midrule
1 & 79.8 & 79.0 & $+0.8$ \\
2 & 77.2 & 75.5 & $+1.7$ \\
3 & 78.5 & 77.2 & $+1.3$ \\
4 & 80.0 & 78.8 & $+1.2$ \\
5 & 79.5 & 79.0 & $+0.5$ \\
\midrule
Mean $\pm$ SD & $79.0 \pm 1.2$ & $77.9 \pm 1.5$ & $+1.1$ \\
\bottomrule
\end{tabular}
\end{table}

The repeated runs quantify training variation and test whether the close anchor ordering is a seed accident.

\paragraph{Run-to-run variation is modest but non-zero.} The \textsc{EffVLA} score varies by
$1.2$ points (SD) across seeds and the \textsc{OFT} score by $1.5$. The $\sim\!1$-point threshold
used elsewhere describes evaluation noise, whereas this table directly measures training-seed
variation for the two models in this comparison.

\paragraph{The \textsc{EffVLA}--\textsc{OFT} ordering is stable.} \textsc{EffVLA} leads in all
five seeds, by $+0.5$ to $+1.7$ points (mean $+1.1$). This is the purpose of the repeated-seed
check: although the absolute scores fluctuate, the single-run ordering of this close comparison
is reproduced. We report it as a small and consistent advantage rather than a headline result.

\section{Cross-Configuration Initialization Check}
\label{app:pairs}

A multi-seed study of all $63$ cells was out of reach; Appendix~\ref{app:seeds} instead checks
the stability of the close \textsc{EffVLA}--\textsc{OFT} ordering. Across the broader sweep,
we can separately report how often the initialization effect reappears when configuration choices change.
Table~\ref{tab:init-pairs} lists every L1-regression pair of cells in the study that differ \emph{only} in the
action head's initialization, holding the decoder, parameter count, latency, module scales and
inference budget fixed.

\begin{table}[htbp]
\centering
\caption{\textbf{Every L1 initialization-matched pair in the study.}  LIBERO-Plus mean success
rate (\%).  Seven rows are module-scale configurations, one repeats the anchor at four inference passes, one is the larger $L\!=\!g$ probe, and the last is a second benchmark and embodiment.}
\label{tab:init-pairs}
\begin{tabular}{lccc}
\toprule
$V/L/A$ & random init & \textsc{VLM-init} & $\Delta$ \\
\midrule
$l/l/l$ (1 pass)  & 72.7 & 79.8 & $+7.1$ \\
$l/l/m$           & 70.2 & 78.0 & $+7.8$ \\
$l/l/s$           & 69.6 & 70.4 & $+0.8$ \\
$l/m/l$           & 69.9 & 76.8 & $+6.9$ \\
$l/s/l$           & 73.0 & 78.1 & $+5.1$ \\
$l/g/l$           & 73.9 & 77.8 & $+3.9$ \\
$m/l/l$           & 69.5 & 76.8 & $+7.3$ \\
$s/l/l$           & 69.2 & 76.4 & $+7.2$ \\
$l/l/l$ (4 passes) & 74.8 & 79.6 & $+4.8$ \\
\midrule
SimplerEnv (Bridge, WidowX) & 54.2 & 61.5 & $+7.3$ \\
\bottomrule
\end{tabular}
\end{table}

All ten differences are positive, with a mean of $+5.8$ points. We treat this as descriptive
robustness across configurations, not as ten independent trials: the pairs reuse the same data,
model components, training pipeline, and (for the vision and language sweeps) the same $l/l/l$
anchor. A sign-test $p$-value would therefore rely on an unestablished independence assumption
and is not reported. The one near-zero case, $+0.8$ at $A\!=\!s$, is the smallest
action head in the sweep, which is what the account in Section~\ref{sec:mechanism} predicts: a
head with few layers has little capacity to inherit into.

We state plainly what this does and does not bound. The positive direction recurs while changing
the vision encoder, language backbone, action-head size, inference budget, benchmark, and
embodiment. Correlation across these configurations means the table neither supplies ten
independent replications nor bounds the run-to-run variance of the initialization contrast.
The priority statistical follow-up is a paired multi-seed comparison of \textsc{EffVLA} with
the architecture-matched \textsc{OFT-TF} baseline.

\section{Real-Robot Evaluation Protocol}
\label{app:realrobot}

\paragraph{Platform and data.}  We instantiate \textsc{EffVLA} unchanged on the SO-ARM101 low-cost
open-source 6-DOF arm: the same $4$-layer transformer-block head copied from the last four layers of
Qwen2.5-3B, single-pass L1, read through the same KV-share interface, with no axis of
Section~\ref{sec:exploration} re-tuned for the new embodiment.  Only the data pipeline changes.
Stage~A action-head pretraining uses the SO-ARM101 community's open-source teleoperation corpus in
place of DROID, and Stage~B fine-tunes on demonstrations we collected by teleoperating the arm on the five target
tasks, merged into one training set: a single model is evaluated on all five, not one model per task.

\paragraph{Task suite.}  The workspace holds three objects, a yellow cube, a blue marker, and a
blackboard eraser, together with three compartments of increasing size.  Each task is issued as a single
natural-language instruction naming every object and the compartment it belongs in, so difficulty is set
by how many objects must be handled in one instruction rather than by the difficulty of any individual
grasp.  Table~\ref{tab:realrobot} gives the five instructions verbatim.

The multi-object tasks are the informative ones.  A single instruction names two or three
object--compartment pairs, so the policy has to track which pairs it has already satisfied and re-ground
the instruction on the remaining ones; an error early in the sequence is not recoverable and the episode
is scored as a failure.

\begin{table}[htbp]
\centering
\caption{\textbf{\textsc{EffVLA} on SO-ARM101.}  The five instructions exactly as issued to the
policy, with ten trials each.  Success requires every object named in the instruction to end in the
compartment the instruction names, so an error early in a multi-object sequence fails the episode.}
\label{tab:realrobot}
\begin{tabular}{clp{0.60\linewidth}c}
\toprule
\# & Objects & Instruction & Success \\
\midrule
1 & 1 & \emph{Pick up the yellow cube and place it in the small compartment.} & 10/10 \\
2 & 1 & \emph{Pick up the blue marker and place it in the medium compartment.} & \phantom{0}9/10 \\
\midrule
3 & 2 & \emph{Pick up the blackboard eraser and place it in the large compartment, then pick up the yellow cube and place it in the small compartment.} & \phantom{0}9/10 \\
4 & 2 & \emph{Pick up the blue marker and place it in the medium compartment, and pick up the yellow cube and place it in the small compartment.} & \phantom{0}7/10 \\
\midrule
5 & 3 & \emph{Pick up the blackboard eraser and place it in the large compartment, pick up the blue marker and place it in the medium compartment, and pick up the yellow cube and place it in the small compartment.} & \phantom{0}5/10 \\
\midrule
\multicolumn{3}{l}{\textbf{Total}} & \textbf{40/50} \\
\bottomrule
\end{tabular}
\end{table}

Success falls monotonically with the number of objects that must be handled in sequence,
from $10/10$ and $9/10$ on the single-object tasks to $5/10$ on the three-object one.  This is the
expected shape for a policy executed open-loop over a long horizon: the per-object success rate
stays high, but the episode requires every stage to succeed, so errors compound.

\paragraph{What this experiment does and does not establish.}  We report the real-robot trials as a
feasibility and transfer result, not a comparative one.  We do not train and evaluate the alternative
action heads (\textsc{OFT}, a randomly initialized transformer head, or a flow-matching head) on the
same hardware, so these trials show that the recipe selected on LIBERO carries over to a different and
much cheaper embodiment with the design axes unchanged; they do not establish that it remains the
preferred configuration in the physical setting.  A like-for-like on-hardware comparison is the natural
next step and is the evidence a stronger deployment claim would need.

\section{Additional Attention Rollouts}
\label{app:attention}

This appendix supports the aggregate instruction-attention result of Section~\ref{sec:mechanism} (Figure~\ref{fig:attention-summary}) with the token-level detail and four further single rollouts.

\paragraph{Token-level detail.} We measure attention at the action head's first layer (L0, the layer nearest the backbone) over $128$ rollouts, $64$ per condition, spanning the four LIBERO-Plus task suites at the $V\!=\!L\!=\!A\!=\!l$ cells. The word ranking in Figure~\ref{fig:attention-summary}b covers instruction words only: the chat-template role tokens that wrap the prompt are not part of the instruction and are excluded. They are nevertheless prominent in the full text-token distribution: the \texttt{assistant} role token alone attracts the single largest share of the \textsc{VLM-init} head's text attention ($0.096$ mean per-token mass, against $0.005$ under random initialisation), more than twice the highest-ranked instruction word. This may reflect an attention-sink or prompt-boundary effect, so the model's learned saliency need not match a human noun-centered reading. The broader pattern remains clearer and more concentrated under \textsc{VLM-init}, while the aggregate instruction-word analysis separately shows elevated attention on task-defining object and location nouns.

\begin{figure}[p]
\centering
\includegraphics[width=\linewidth]{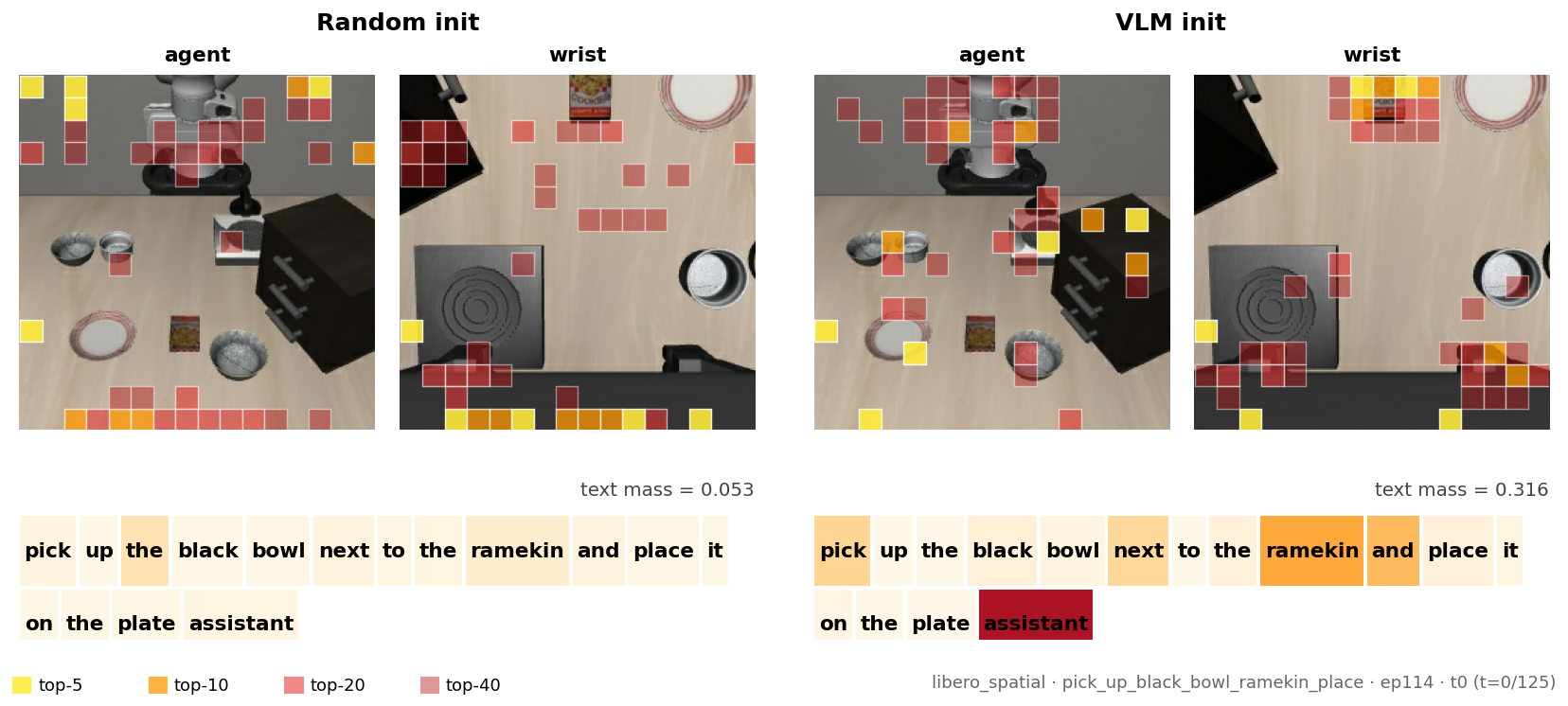}\\[6pt]
\includegraphics[width=\linewidth]{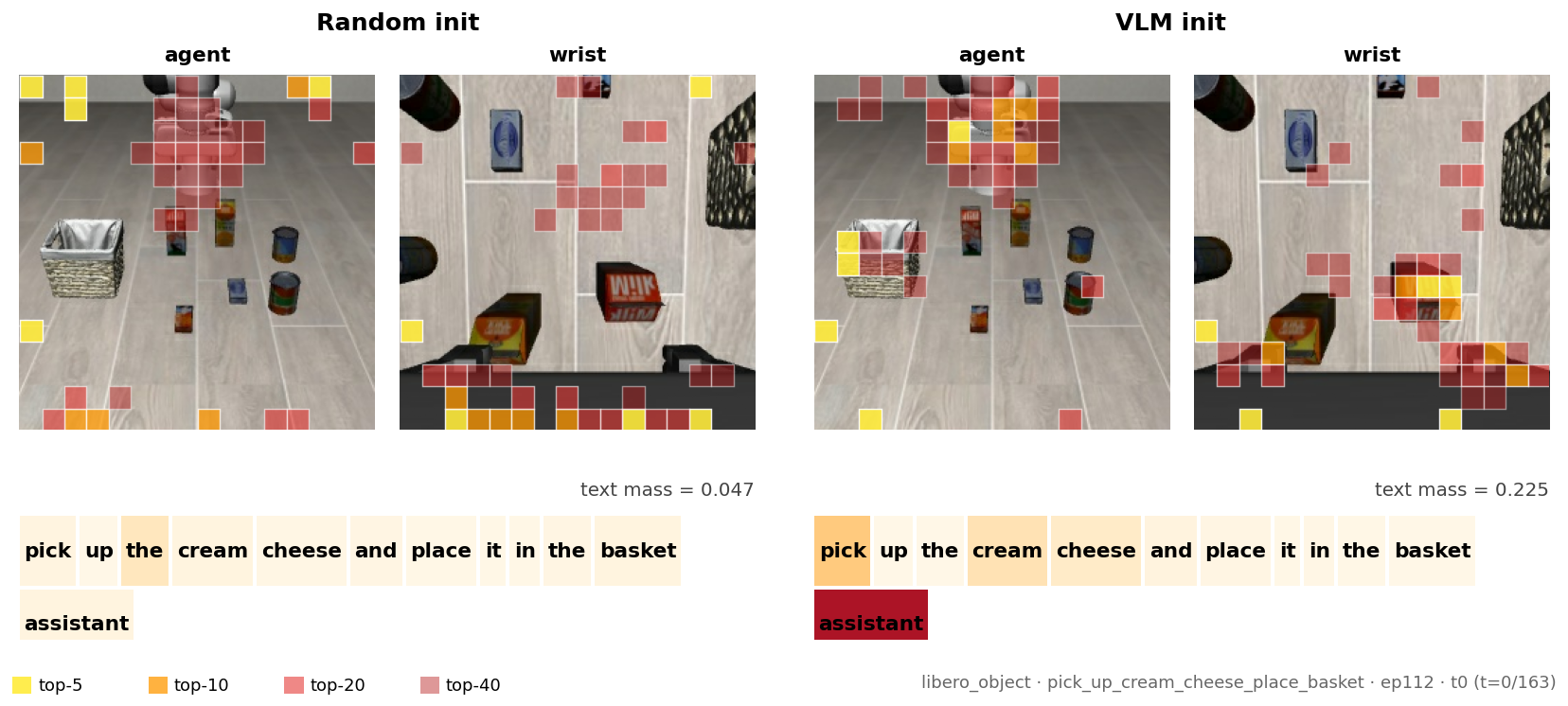}
\caption{\textbf{Additional attention rollouts (spatial and object suites).} Same format as Figure~\ref{fig:attention-case}: random init on the left, \textsc{VLM-init} on the right; vision panels overlay the top-$K$ attended patches (agent and wrist views), and the ribbon below each shades every instruction token by its attention.}
\label{fig:attention-cases-a}
\end{figure}

\begin{figure}[p]
\centering
\includegraphics[width=\linewidth]{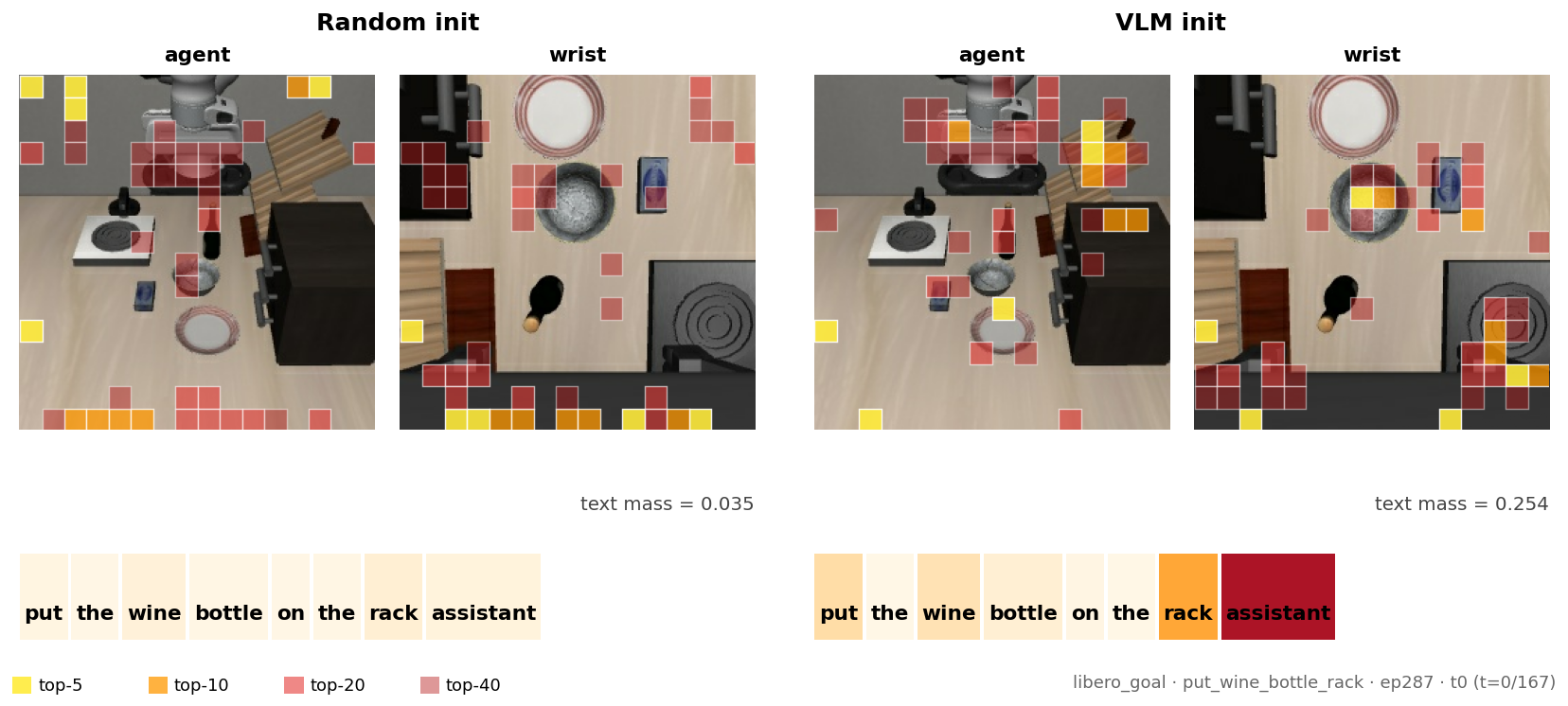}\\[6pt]
\includegraphics[width=\linewidth]{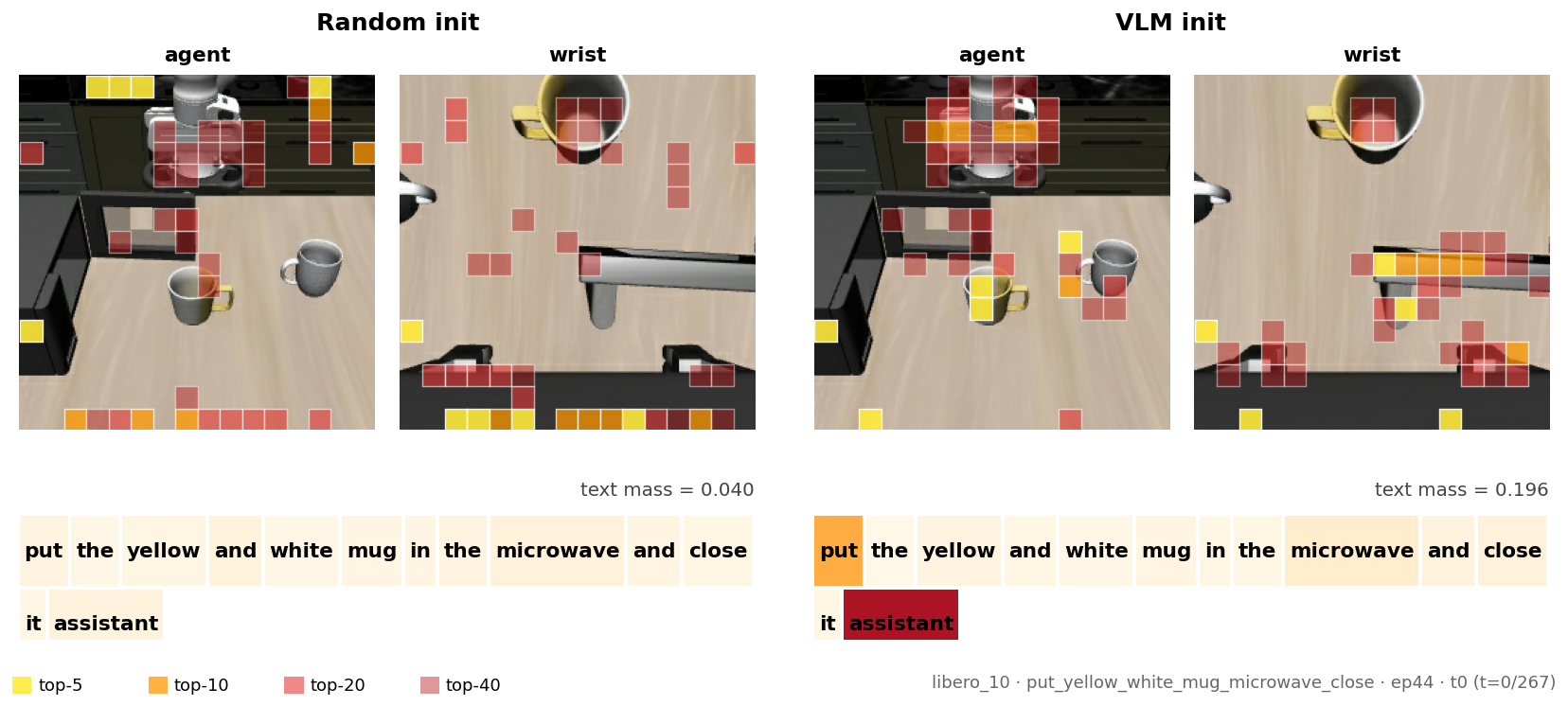}
\caption{\textbf{Additional attention rollouts (goal and long suites).} Format as in Figure~\ref{fig:attention-cases-a}.}
\label{fig:attention-cases-b}
\end{figure}

\paragraph{Per-rollout examples.} Figures~\ref{fig:attention-cases-a} and~\ref{fig:attention-cases-b} show four further single-rollout examples in the same format as Figure~\ref{fig:attention-case}, one from each LIBERO-Plus task suite. In every case the \textsc{VLM-init} head places several times more mass on the instruction than the random-initialised head and concentrates it on the object and location nouns that name the task (\emph{bowl}, \emph{ramekin}, \emph{cream cheese}, \emph{wine bottle}, \emph{rack}, \emph{mug}, \emph{microwave}), while the random head leaves the instruction almost unattended and spreads its vision attention diffusely across the scene.

\section{Weight-Space Alignment: Additional Detail}
\label{app:repalign}

This appendix expands the CKA analysis of Figure~\ref{fig:cka3} with the per-module breakdown. Because the $L_1$ transformer-block head is architecturally identical to a backbone block, its projections share the backbone's shape, so the CKA between each action-head projection and the corresponding backbone layer is well-defined. Read per module on the $V\!=\!L\!=\!A\!=\!l$ cells, the K$+$V ``backbone-information'' and output projections are preserved most (CKA $0.79$ and $0.81$) while the query path Q adapts most (CKA $0.54$), consistent with the head keeping \emph{what} the backbone reads while re-purposing \emph{how} it attends, the weight-space counterpart of the instruction-attention behaviour in the main text (Figure~\ref{fig:attention-case}).

We read this evidence as correlational. It accounts for the attention behaviour and the success-rate pattern, and it is what motivates reading the whole design space through alignment, but we manipulate alignment only through initialization, so it does not establish that alignment, rather than these particular pretrained weights, is the operative variable (a K-rank truncation probe was inconclusive under the LIBERO-Plus ceiling). The decisive control would initialize the head from the same backbone weights with the alignment destroyed, by permuting the layers or applying a random orthogonal rotation per matrix, holding the weight statistics fixed: if alignment is what matters, both the CKA and the $+7.1$ gain should collapse to the random-init level. It is consistent with the cross-module transfer results of \citet{kumar2022fine,lee2022surgical}.

\section{Limitations and Extended Discussion}
\label{app:limitations}

This appendix expands the main-text Limitations section (Section~\ref{sec:limitations}) with the supporting evidence, the failure modes we observe, and concrete steps that would extend the claims.

\paragraph{Why action-head scaling depends on initialisation (Finding~2).} Section~\ref{sec:scaling} shows that action-head scaling returns about $4$ success points per ms when the head inherits the language backbone's weights, and roughly a third as much when it does not. The interpretation that fits the numbers is that copied transformer blocks begin from a more compatible parameter structure, allowing added capacity to be used more effectively. This is consistent with the transfer-learning results of \citet{kumar2022fine,lee2022surgical} discussed in Section~\ref{sec:related}. Appendix~\ref{app:repalign} supplies weight-space evidence (final Weight-Gram CKA $0.76$ versus $0.24$) and a separate attention analysis shows greater instruction-token mass, but neither establishes activation-level representation alignment or causality. Whether the same coupling extends to other modular pretrained systems is also not something we test here.

\paragraph{Backbone-family specificity.}  Every cell in our experiment uses SigLIP2 as the vision encoder and Qwen2.5 as the language backbone, so the $+7.1$-point initialisation gain rests on the assumption that Qwen2.5's last transformer layers are useful when used as action-head initial conditions.  Other VLM families differ in normalisation (RMS vs.\ LayerNorm), positional encoding (RoPE vs.\ ALiBi vs.\ sinusoidal), and attention pattern (GQA vs.\ MHA vs.\ MLA), and any of these could change the size or even the sign of the initialisation effect.  The most direct way to test this is a replication on a Llama-3- or Gemma-2-based backbone using the same factorial; we treat the present experiment as the controlled measurement on one backbone family rather than a universal claim.

\paragraph{Benchmark scope and a measured failure mode.}  LIBERO-Plus is simulated and emphasises short-horizon manipulation; the saturation past $l$ that we read off Figures~\ref{fig:module-scaling}--\ref{fig:module-budget} is therefore a saturation on \emph{this} benchmark.  Two concrete failure modes are visible already.  First, on the \texttt{Sensor Noise} perturbation axis the plain MLP head (\textsc{OFT}) beats \textsc{EffVLA} by $13$~points (Table~\ref{tab:per-axis-ablation}, $79.4$ vs.\ $66.1$).  Second, the per-task-suite breakdown (Figure~\ref{fig:scaling-per-suite}) shows that \texttt{long-10} has the steepest substrate-specific rises and is the suite least saturated by $l$.  We expect harder, longer-horizon, or higher-precision benchmarks (e.g.\ RoboCasa, mobile manipulation, real long-horizon kitchen tasks) to re-open the case for larger models; a replication on at least one such benchmark is the natural next step.

\paragraph{Statistical resolution.} Each of the $63$ cells in the sweep uses a single random seed. The five-seed check at $l/l/l$ directly addresses the close \textsc{EffVLA}--\textsc{OFT} comparison: run-to-run standard deviations are $1.2$--$1.5$ points, yet \textsc{EffVLA} ranks higher in all five runs (Appendix~\ref{app:seeds}). This shows that the reported single-run ordering for that comparison is stable despite modest score variation. We still treat sub-point gaps elsewhere as unresolved and do not claim ordinal rankings on harder perturbation axes where multiple cells sit within $1$--$3$ points.

\paragraph{Architectural caveat: the KV-share interface.}  All four transformer-block substrates share the same VLM--policy interface: the action stack reads the language backbone's KV cache directly.  This interface is what makes the $L\!\leftrightarrow\!A$ depth--width coupling of Appendix~\ref{app:scaling} unavoidable and what makes \textsc{VLM-init} architecturally clean (since the head and the backbone have identical block shape).  We did not ablate the interface itself.  Whether the rules survive under a Q-former, cross-attention, or learned-query connector is an open question, and the most informative replacement experiment would compare KV-share against one alternative connector while holding everything else fixed.

\paragraph{Hardware specificity of the latency analysis.}  All wall-clock latencies are measured on a single NVIDIA RTX~5090 in bf16 with batch size $1$.  On edge hardware (Jetson Orin, mobile NPUs) the per-module latency ratios shift because vision encoders and language backbones have different memory-bandwidth profiles, and the action-head dominance of Finding~2 may compress or invert.  A multi-platform latency replication (a server GPU, an edge GPU, a mobile chip) would establish whether the recipe's Pareto-optimality is platform-independent.

\paragraph{Real-robot evidence.}  The SO-ARM101 deployment validates feasibility without architectural change on five language-specified sorting tasks ($10$ trials each, $50$ total) using one inexpensive 6-DOF arm. These tasks belong to one task family and use one embodiment, and no alternative action head is evaluated on hardware. We therefore do not claim generalisation to other task families or to bimanual, mobile, or high-precision platforms (e.g.\ Franka, Aloha, mobile-manipulator stacks); a broader comparative evaluation is the most consequential gap to close.

\paragraph{Fixed inputs.}  We hold the input image resolution at $256$\,px and the action chunk size at $8$ throughout.  Both affect the action-head's task in non-trivial ways: a longer chunk lengthens the parallel decode the language-initialised head is performing, and a higher resolution changes the visual-token budget the vision encoder produces.  A sweep over $\{4, 8, 16\}$ chunk sizes on \textsc{EffVLA} would be the cleanest stress test of Finding~2, since action-head scaling is what we predict should benefit most.

\end{document}